\documentclass[final]{siamltex}
\usepackage{algorithm,algpseudocode}
\usepackage{latexsym, graphicx, epsfig, amsmath, amsfonts,amssymb,bm}
\usepackage{caption, subcaption}
\usepackage{epstopdf}
\usepackage{booktabs}
\usepackage{array}
\usepackage{color}
\usepackage{lineno}
\usepackage{url}

\def\be{\begin{equation}}
\def\ee{\end{equation}}

\def\x{\mathbf{x}}

\def\f{\mathbf{f}}

\def\PPhi{\boldsymbol{\Phi}}

\def\N{\mathbf{N}}

\def\A{\mathbf{A}}
\def\b{\mathbf{b}}

\def\R{{\mathbb R}}

\newcommand{\argmin}{\operatornamewithlimits{argmin}}

\newcommand{\figref}[1]{Fig.~\ref{#1}}

\title{On Generalized Residue Network for Deep Learning of Unknown Dynamical Systems}

\author{Zhen Chen \and Dongbin
       Xiu\thanks{Department of Mathematics,
		The Ohio State University, Columbus, OH 43210, USA.
		{\tt chen.7168@osu.edu, xiu.16@osu.edu.}
		Funding: This work was partially supported by AFOSR FA9550-18-1-0102.}
}

\begin{document}
\maketitle
\begin{abstract}
We present a general numerical approach for learning unknown dynamical systems using deep neural networks (DNNs). Our method
is built upon recent studies that identified residue network (ResNet) as an effective neural network structure. 
In this paper, we present a generalized ResNet framework and broadly define "residue" as the discrepancy between observation data and prediction made by another model,  which can be an existing coarse model or reduced order model. In this case, the generalized ResNet serves as a model correction to the existing model and recovers the unresolved dynamics. When an existing coarse model is not available, we present numerical strategies for fast creation of coarse models, to be used in conjunction
with the generalized ResNet. These coarse models are constructed using the same data set and thus do not require additional resource.
The generalized ResNet is capable of learning the underlying unknown equations and producing predictions with accuracy higher than the standard
ResNet structure. This is demonstrated via several numerical examples, including long-term prediction of a chaotic system.
\end{abstract}

\section{Introduction} \label{sec:intro}
 
There has been a surge of interest in developing algorithms to recover unknown governing equations via observation data.
Most efforts focus on recovering unknown system of ordinary differential equations, i.e., dynamical systems.
Some (relatively) earlier efforts seek to construct certain optimal linear approximation for the unknown system. These include dynamic model decomposition (DMD)  (\cite{Schmid_2010}) and its variants for nonlinear     
systems using Koopman theory, cf., \cite{brunton2017chaos, Kutz_2014, kutz2016dynamic}.
More recent efforts cast the problem into an approximation problem, 
where the unknown governing equation is treated as
a target function relating the data of the state variables to their temporal
derivatives. Methods along this line of approach usually seek
exact recovery of the equations by using certain sparse approximation 
techniques (e.g., \cite{tibshirani1996regression}) from a large set of dictionaries; see, for example, \cite{brunton2016discovering}. 
Studies have  been conducted to deal with noises in data
\cite{brunton2016discovering, schaeffer2017sparse,kang2019ident}, corruptions in data
\cite{tran2017exact}, limited data \cite{schaeffer2017extracting}, 
partial differential equations \cite{rudy2017data,
	schaeffer2017learning}, etc.
Variations of the approaches have been developed in conjunction with other
methods such as
model selection approach
\cite{Mangan20170009}, Koopman theory \cite{brunton2017chaos}, 
Gaussian process regression \cite{raissi2017machine,RAISSI2018125},
and expectation-maximization approach \cite{nguyen2019like}, to name a
few. 
Methods using standard basis functions and without requiring exact
recovery were also  developed for dynamical systems
\cite{WuXiu_JCPEQ18}  and Hamiltonian systems \cite{WuQinXiu2019}.

The use of modern machine learning techniques, particularly deep neural networks (DNNs), offers a new line of approaches for the task.
DNN structures have been developed to recover
ordinary differential equations (ODEs) \cite{raissi2018multistep,qin2018data,rudy2018deep} and partial differential equations (PDEs)
\cite{long2017pde,raissi2017physics1,raissi2017physics2,raissi2018deep,long2018pde,sun2019neupde}. 
It was shown that residual network
(ResNet) is particularly suitable for equation recovery, in the sense
that it can be an exact integrator \cite{qin2018data}. 
Neural networks have also been explored for other
aspects of scientific computing, including 
reduced order modeling \cite{HESTHAVEN201855,Pawar_2019}, solution of
conservation laws \cite{RAY2018166,WANG2019289}, multiphase flow simulation \cite{wang2019efficient}, 
high-dimensional PDEs 
\cite{Han8505, Khoo2018}, 
uncertainty quantification \cite{Chan_2018,Tripathy_2018, Zhu_2018,karumuri2019simulator}, etc.

The focus and contribution of this paper is on generalization of ResNet (gResNet) structure for recovering unknown dynamical systems. In the standard ResNet  method 
for equation recovery, the residue is defined as the difference between the data inputs and data outputs, and a deep neural network is used to model the residue. (For
more detail, see \cite{qin2018data}.) In this paper, we broaden the concept of residue and broadly define it as the difference between the data outputs and {\em the predictive outputs made by another model}, which
shall be referred to as ``prior prediction'' hereafter.
We then use a standard feedforward neural network to model this generalized residue and to construct the final prediction, hereafter referred to as ``posteriori prediction'',  by taking into account of the prior prediction. 
The gResNet structure can then be viewed as a model correction method for the prior predictive model. The prior model can be an existing coarse model, reduced order model, empirical model, etc.
The deep neural network in the gResNet is used to construct a governing equation for the discrepancy between the prior model and the true model. We remark that model correction/calibration
is an ongoing research topic, where several methods exist. See, for example, \cite{BayarriEtAl_07, HeXiu_JCP16, HigdonEtAl_04, JosephM_09, KennedyOHagan01, SargsyanNG_15, WangCT_09,
  QianWu_08} and the references therein. Our method here represents a new and drastically different approach, via the use of deep neural networks, to this line of study.
It is also straightforward to see that the generalized ResNet (gResNet) includes the standard ResNet as a special case, in the sense that in the standard ResNet
the prior model takes the trivial form of an identity operator and the prior prediction is the same as the data inputs.

In many practical situations, one may not possess a prior model and only has access to data. In this case, we propose a number of numerical strategies to create a prior model
using the same data set. In particular, we discuss two approaches. The first one seeks to construct an affine approximation for the underlying system as the prior model. This is an extension of the well known DMD (dynamic mode decomposition) method, which utilizes linear approximation (\cite{Schmid_2010}).
The use of affine approximation, which has a constant vector term in addition to linear transformation, allows one to model possible non-homogenenous terms in the underlying dynamical system.
This proves to be more flexible in practice.
The other method seeks to construct the prior model as a nonlinear approximation using a single hidden layer NN.  This is an extension of the affine approximation modeling, which is a linear procedure.
Both approaches can produce prior models in efficient manner and do not increase the overall computational cost of the entire modeling process. Note that
the emphasis of the paper is on the gResNet framework, and the discussion of the two approaches for prior model construction is to provide some viable choices. In practice, one is free to choose any other suitable (for the given problem) method to construct the prior model.


\section{Setup and Preliminaries} \label{sec:setup}

Let us consider an autonomous dynamical system
\be \label{eq:govern}
\frac{d\x}{dt} = \f(\x), \qquad \x(t_0) = \x_0,
\ee
where $\x\in\R^n$ are the state variables. Note that the dimension $n$
can be large, especially when the system is obtained via a spatial
discretization of a PDE system.
In this paper, we assume  the form of the governing equations
$\f:\R^n\to\R^n$ is unknown. What is available is measurement data of
the solution states. Let $t_0 < t_1 < \cdots< t_K$ be a sequence of time
instances. We use
\be \label{raw_data}
\x^{(i)}_k = \x(t_k; \x_0^{(i)}, t_0), \qquad k= 1, \dots,K, \quad i=1,\dots, I_{traj},
\ee
to stand for the solution state at time instance $t_k$, originated from the
$i$-th initial state $\x_0^{(i)}$ at $t_0$, for a total number $I_{traj}$
trajectories.
Note that the data can also contain
measurement noises, which are usually modeled as random variables.
Our goal is then to
create an accurate approximation model for the unknown governing
equations by using the solution state data.

\subsection{Flow Map and Data Pairing}

While many of the existing work seeks to construct a system $d\x/dt =
\widetilde{f}(\x)$ as an approximation to the unknown system
\eqref{eq:govern}, we here adopt a different approach developed in
\cite{qin2018data}. This approach does not directly approximate the
right-hand-side of the system \eqref{eq:govern}. Instead, it seeks to
approximate the flow map of the underlying system.

Let $\PPhi_s: \R^n\to \R^n$ be the flow map of system
\eqref{eq:govern}, whose solution follows 
\be
\x(t; \x_0, t_0) = \PPhi_{t-t_0}(\x_0).
\ee
Note that for autonomous systems the time variable $t$ can be
arbitrarily shifted and only the time difference, or time lag, $t-t_0$ is
relevant. Also, $\PPhi_{s+r} =  \PPhi_s \circ \PPhi_r$.
The flow map completely determines the evolution of the solution from
one state to another state at a different time. Recovery of the flow map allows
one to conduct prediction of the system via recursive use of the flow
map. Let $\Delta_k = t_k - t_{k-1}$, $k=1,\dots$ be the time
differences in the time instances. We then organize the solution
state data in \eqref{raw_data} into pairs, separated by time lag $\Delta_k$,
\be \label{data_pairs}
\left\{\x^{(i)}_{k-1}, \x^{(i)}_{k}\right\}, \qquad k=1,\dots, K, \quad
i=1,\dots, I_{traj}.
\ee
For notational convenience, hereafter we
assume $\Delta_k = \Delta$ is a constant for all $k$.
We then denote the entire data set as
\be \label{set}
\mathcal{S} = \{ (\x_j^{(1)}, \x_j^{(2)}): j=1,\dots, J\},
\ee
where $J=K\times I_{traj}$ is the total number of data pairs. For each
$j$-th pair, $\x^{(1)}_j$ is the ``initial state'',
$\x^{(2)}_j$ is the ``end state'', and the two states are separated by
the time lag $\Delta$. In the noiseless case, the two states are
governed by the (unknown) flow map such that
\be \label{Phi}
\x^{(2)}_j = \PPhi_\Delta (\x_j^{(1)}).
\ee

\subsection{ResNet Modeling of Flow Map}

In \cite{qin2018data}, a method was proposed to discover
the unknown dynamical system \eqref{eq:govern} via numerically
approximating its underlying flow map. This is accomplished by using
the data paris \eqref{set} to approximate the unknown flow map over
discrete time step $\Delta$ in \eqref{Phi}. Once this $\Delta$-flow
map
is constructed, it can be recursively applied to conduct solution
prediction over much longer time horizon.

Moreover, the work of
\cite{qin2018data} also proposed to utilize residue network (ResNet)
to conduct deep learning of the $\Delta$-flow map.
While the standard feedforward deep neural networks (DNNs) utilize
multiple hidden layers to approximate input-output maps, ResNet
applies the identity operator on the input data and superimposes it on
the neural network outputs. This effectively creates a DNN modeling
for the ``residue'' of the input-output data.

Let ${\mathcal{N}}(\cdot;\Theta): \R^n\to \R^n$ be the operator from a standard fully
connected feedforward neural network, where $\Theta$ denotes its
parameter set. Its corresponding ResNet then creates a mapping
\be \label{ResNet_map}
\N(\cdot;\Theta) = \mathcal{I} + {\mathcal{N}}(\cdot;\Theta),
\ee
where $\mathcal{I}$ is the identity operator. By applying this structure
to the data set \eqref{set} and defining a loss function in the
form of
\be \label{loss:ResNet}
L(\Theta) = \frac{1}{J} \sum_{j=1}^J \left\|\x^{(2)}_j - \N(\x^{(1)}_j;\Theta)\right\|^2,
\ee
one can train a ResNet model
\be 
\x_j^{(2)}\approx \x_j^{(1)} + \mathcal{N}(\x_j^{(1)}; \Theta^*),
\ee
where $\Theta^*$ is the network parameter set after successful
training. Upon obtaining the trained network model, one can
recursively apply the model to conduct system prediction in the
following form
\be \label{ResNet_model}
\x_{k+1} = \x_{k} + \mathcal{N}(\x_{k}; \Theta^*), \qquad k=0,1,\dots,
\ee
for a given initial condition $\x_0$.
Even though
the form of \eqref{ResNet_model} resembles Euler forward time stepper,
it was shown in \cite{qin2018data} that this model is an exact time
integrator, in the sense that there is no error associated with the
time step $\Delta$.

\section{Generalized ResNet (gResNet) Modeling} \label{sec:gResNet}

In this section we present a generalization of the ResNet (gResNet) 
method developed in \cite{qin2018data} for approximating unknown equations. We first
present the general approach of the gResNet method and then discuss a
few practical options.

\subsection{General Approach of gResNet}

Let $\mathcal{L}: \R^n\to \R^n$ be an operator such that
\be \label{L}
\x(t_{k+1}) \approx \mathcal{L}(\x(t_k)).
\ee
The proposed gResNet mapping takes the following form
\be \label{gResNet_map}
\N(\cdot;\Theta) = \mathcal{L} + {\mathcal{N}}(\cdot;\Theta),
\ee
where ${\mathcal{N}}(\cdot;\Theta)$ is the operator associated with a
standard fully connected feedforward neural network.
The training of the gResNet is similar to that of ResNet. With the
data set \eqref{set} and loss function
\be \label{loss_gResNet}
L(\Theta) = \frac{1}{J} \sum_{j=1}^J \left\|\x^{(2)}_j -
  \N(\x^{(1)}_j;\Theta)\right\|^2= \frac{1}{J} \sum_{j=1}^J \left\|\x^{(2)}_j -\mathcal{L}(\x_j^{(1)})-
  \mathcal{N}(\x^{(1)}_j;\Theta)\right\|^2,
\ee
one can train for the optimized network parameter set $\Theta^*$ and
obtain the corresponding trained network $\mathcal{N}(\cdot,\Theta^*)$.
Subsequently, we obtain the corresponding gResNet prediction model
\be \label{eq:gResNet}
\x_{k+1} = \mathcal{L}(\x_k) + \mathcal{N}(\x_k;\Theta^*), \qquad k=0,1,\cdots.
\ee

It is straightforward to see, upon comparing \eqref{gResNet_map} with
the standard ResNet mapping
\eqref{ResNet_map}, that the original ResNet model is a special case
of the gResNet when $\mathcal{L} = \mathcal{I}$, the identity operator.
In fact, when the time lag $\Delta$ is sufficiently small, $\x_{k+1} =
\x_k +\mathcal{O}(\Delta)$ and the exact
flow map satisfies
$$
\PPhi = \mathcal{I} + \mathcal{O}(\Delta).
$$
This implies that the identity operator is a reasonable choice for
$\mathcal{L}$ in term of \eqref{L}.
This is the reason why the original ResNet provides a computational
advantage, as discussed in \cite{qin2018data}. When $\Delta$ is not
sufficiently small, the standard ResNet may not be advantageous.

\subsection{Model Correction for Known $\mathcal{L}$}
\label{sec:L}

The operator $\mathcal{L}$ \eqref{L} in the gResNet model \eqref{gResNet_map} is
critical. It should be available prior to the gResNet model
construction. Hereafter we will loosely refer the operator $\mathcal{L}$ to as ``prior
model''. In general, $\mathcal{L}$ stands for any available
models for the underlying dynamical system. This can be a linear
approximation, a reduce order model, a coarse grained model, etc. By
construction \eqref{gResNet_map}, the fully connected hidden layers inside
gResNet correspond to the operator $\mathcal{N}$
and are used to approximate a generalized residue in the following sense
$$
\mathcal{N}(\cdot;\Theta^*) \approx \PPhi - \mathcal{L},
$$
where $\PPhi$ is the flow map of the true model. If one assumes
that the
operator $\mathcal{L}$ is an approximation of the underlying dynamics
via \eqref{L} such that
$$
\x(t_{k+1}) = \mathcal{L}(\x(t_k)) + \mathcal{O}(\epsilon),
$$
then it is natural to see the neural network operator
$\mathcal{N}\sim\mathcal{O}(\epsilon)$. Consequently, this provides a
computational advantage for the gResNet.

Since the prior model $\mathcal{L}$ represents a ``coarse'' approximation of the
underlying dynamics, one can then view the network operator
$\mathcal{N}$ in the gResNet as a ``model correction'' to
$\mathcal{L}$, as shown in \eqref{gResNet_map}. And we will refer to
the trained gResNet model \eqref{gResNet_map} as ``posteriori model'' hereafter.
Examples of coarse, or reduced order, modeling are abundant in
scientific computing. Here we give a very specific example, which is adopted
from \cite{PavliotisStuart_2008} and will be used in this paper as a
numerical test.

Consider a true (and unknown) dynamical system
\be \label{ODE4}
\left\{
\begin{split}
  \frac{dx_1}{dt} &= -x_2-x_3, \\
  \frac{dx_2}{dt} &= x_1+\frac{1}{5}x_2, \\
  \frac{dx_3}{dt} &= \frac{1}{5} + y - 5x_3,\\
  \frac{dy}{dt} &= -\frac{y}{\epsilon} + \frac{x_1x_3}{\epsilon},
\end{split}
\right.
\ee
where $\epsilon>0$ is a real parameter. This is a chaotic system. A reduced order model for this
system is
\be \label{ODE3}
\left\{
\begin{split}
    \frac{dX_1}{dt} &= -X_2-X_3, \\
  \frac{dX_2}{dt} &= X_1+\frac{1}{5}X_2, \\
  \frac{dX_3}{dt} &= \frac{1}{5} + X_3(X_1 - 5),
\end{split}
\right.
\ee
where the fast variable $y$ is averaged out. The reduced system
\eqref{ODE3} serves as a good approximation of the true system
\eqref{ODE4} when $\epsilon \ll 1$. This is our prior model in this
case. The operator $\mathcal{L}$ of this prior model
does not have an explicit expression and needs to be computed via
solving \eqref{ODE3} numerically.

\subsection{Affine Approximation for Unknown $\mathcal{L}$}
\label{sec:DMD}

In many practical situations, one does not have an existing prior model and
subsequently the operator $\mathcal{L}$ is not available.
In this case, it is possible to construct a prior model and its
associated operator $\mathcal{L}$
using the same dataset. It is also desirable that such an
construction should be reasonably faster than the neural network
training of the gResNet model, in order not to increase the overall
computational cost.
Any efficient method to create an approximation model using the data
set \eqref{set} can be adopted. Here we present a construction
using affine approximation as a possible choice. This affine
approximation is a modification of dynamic model
decomposition (DMD) method (\cite{Schmid_2010}), which has been used as a linear approximation
model for a variety of problems. 

The idea of DMD is to construct a best-fit
linear dynamical model to approximate the underlying system
based on data. For the given data set \eqref{set}, DMD seeks a linear
flow map $\mathbf{A}\in\R^{n\times n}$ such that
\be \label{DMD}
\x_j^{(2)} \approx \mathbf{A} \x_j^{(1)}, \qquad \forall j=1,\dots, J.
\ee
With sufficient number of data pairs, also known as snapshots, the
matrix $\mathbf{A}$ can be solved in a least squares sense. For more detailed discussion of DMD,
see \cite{kutz2016dynamic}.

The form of DMD \eqref{DMD} makes it effective for homogeneous
systems. In order to cope with potential non-homogeneity of the
underlying dynamics, we employ the following modification
\be \label{mDMD}
\x_j^{(2)} \approx \mathbf{A} \x_j^{(1)} + \mathbf{b}, \qquad \forall j=1,\dots, J,
\ee
where $\mathbf{b}\in\R^n$. Hereafter, we will refer this to as
modified DMD (mDMD) method. This effectively creates an affine mapping
as an approximation of the underlying flow map, i.e.,
\be \label{approx:mDMD}
\PPhi_{\Delta}(\x) \approx \A(\Delta) \x + \b(\Delta),
\ee
where the matrix $\mathbf{A}$ and vector $\mathbf{b}$ are solved via
the following optimization problem
\be \label{opt:mDMD}
(\A, \b) = \argmin_{\substack{\hat{\mathbf{A}} \in \R^{n \times n} \\
    \hat{\mathbf{b}} \in \R^n }} \frac{1}{J}\sum_{j=1}^{J} \left \| \x_j^{(2)} -
  \hat{\mathbf{A}} \x_j^{(1)} - \hat{\mathbf{b}} \right \|^2. 
\ee 
%
To solve the optimization problem, we take the data set \eqref{set}
and write
\be
\mathbf{X}_1:= \begin{bmatrix}		\x_1^{(1)}, \cdots, \x_{J}^{(1)} 
	\end{bmatrix}, \qquad
	\mathbf{X}_2 := \begin{bmatrix}
		\x_1^{(2)}, \cdots, \x_J^{(2)}
	\end{bmatrix}.
\ee
Let $\mathbf{1} :=  [1 \cdots 1]^{T}$ be a vector of size $J \times 1$  and 
\be
	\mathbf{\widetilde{X}}_1 := 
	\begin{bmatrix}
		\mathbf{X}_1 \\
		\mathbf{1}^{T}
	\end{bmatrix}.
\ee
The solution to \eqref{opt:mDMD} is then readily available as
\be
\begin{bmatrix}
		\A, & \mathbf{b} 
	\end{bmatrix}
	= \mathbf{X}_2 {\mathbf{\widetilde{X}}_1}^{\dagger},
\ee
where $\dagger$ stands for matrix pseudo inverse.

We now define the $\mathcal{L}$ operator in the gResNet
\eqref{gResNet_map} as the mDMD model, i.e.,
\be \label{L_mDMD}
\mathcal{L}(\x) = \A \x + \mathbf{b}.
\ee
After training the network operator $\mathcal{N}$ using the loss
function \eqref{loss_gResNet} to obtain the trained parameter set
$\Theta^*$, we obtain the mDMD based gResNet prediction model
\be \label{mDMD_gResNet}
\x_{k+1} = \A\x_k+\mathbf{b} + \mathcal{N}(\x_k;\Theta^*), \qquad k=0,1,\cdots.
\ee

Let
$$
\epsilon^{\textrm{mDMD}}_j =  \x_j^{(2)} -
  {\mathbf{A}} \x_j^{(1)} - {\mathbf{b}}, \qquad j=1,\dots, J,
$$
be the residue of the mDMD model construction from the minimization problem
\eqref{opt:mDMD}. It is then straightforward to see that during the training
of the neural network operator $\mathcal{N}$, its loss function
\eqref{loss_gResNet} can be rewritten as
\be \label{loss_mDMDnet}
L(\Theta) = \frac{1}{J} \sum_{j=1}^J \left\|\epsilon^{\textrm{mDMD}}_j-
  \mathcal{N}(\x^{(1)}_j;\Theta)\right\|^2.
\ee
This further indicates that the neural network operator $\mathcal{N}$
is trained to learn the ``residue'' of the prior model, in this case the mDMD model.

\subsection{Adaptive Nonlinear Approximation for Unknown
  $\mathcal{L}$} \label{sec:adaptiveL}

The mDMD (resp. DMD) in the previous section employs affine
(resp. linear) approximation as the prior model
$\mathcal{L}$, and the model $\mathcal{L}$ remains fixed  for all solution data
\eqref{set}. Upon examining the affine approximation \eqref{L_mDMD},
it is evident that its form $\mathcal{L}(\x) =\A\x+\mathbf{b}$
resembles that of a feedforward neural network with a single hidden
layer, where $\A$ resembles the weight matrix and $\mathbf{b}$
resembles the bias vector. Motivated by this observation, we  propose to use a single-layer
neural network as the prior model $\mathcal{L}$.
Let $\mathcal{N}_{\textrm{prior}}(\cdot;\Theta)$ be the operator
associated with a feedforward neural network with a single hidden
layer. We first train this network using the same data set \eqref{set}
with the loss function
\be
\Theta^*_{\textrm{prior}} 
=\argmin_{\Theta} \frac{1}{J}\sum_{j=1}^{J} \left \| \x_j^{(2)} -
  \mathcal{N}_{\textrm{prior}}(\x^{(1)}_j; \Theta) \right \|^2.
\ee
This in turn gives us the trained prior model
\be
\mathcal{L}(\x) = \mathcal{N}_{\textrm{prior}}(\x;
\Theta^*_{\textrm{prior}}).
\ee
We then construct gResNet network by using this operator $\mathcal{L}$
and solving \eqref{loss_gResNet}. Again, it is straightforward to see
that the loss function is equivalent to
\be
L(\Theta) = \frac{1}{J} \sum_{j=1}^J \left\|\epsilon^{\textrm{prior}}_j-
  \mathcal{N}(\x^{(1)}_j;\Theta)\right\|^2,
\ee
where
$$
\epsilon^{\textrm{prior}}_j =  \x_j^{(2)} -
 \mathcal{N}_{\textrm{prior}}(\x^{(1)}_j; \Theta^*_{\textrm{prior}}), \qquad j=1,\dots, J,
 $$
 is the residue of the training error of the prior model
 $\mathcal{L}$.

 Note that one is free to construct the prior model using a neural
 network with multiple hidden layers. However, with the gResNet
 inherently having a network
 $\mathcal{N}$ with multiple layers, there is no compelling reason to
   introduce multiple layers in the prior model, especially that the
   construction of the prior model should be reasonably fast as not to
   increase the overall cost of the model construction.

\section{Numerical Examples} \label{sec:examples}

In this section we present numerical examples to demonstrate the
efficiency of the proposed methods. For benchmarking purpose, in all
examples the true dynamical models are known. We use the true models to
generate synthetic data and then construct the corresponding gResNet
approximation models. The
gResNet models are then used for system predictions and
compared against the solutions of the true systems.

To generate data set in the form of \eqref{set}, we conduct random
sampling for the ``initial condition'' $\x_j^{(1)}$ and use the true
models to advance time lag $\Delta$ to obtain the corresponding
$\x_j^{(2)}$. It has been established in \cite{WuXiu_JCPEQ18} that
random sampling is more effective for equation recovery
work. Depending on the network structure, the total number of data
pairs in \eqref{set}, $J$, is usually kept at $5\sim 10$ times of the
number of parameters in the network structure. This is to ensure that
the network training does not suffer from overfitting issue. Note
there is no comprehensive theory regarding the sufficient number of
data entries to ensure accurate network training. Therefore, we
purposefully keep the data set sufficiently large so that we can focus
on the fundamental properties of the network.
All models are trained via the loss function \eqref{loss_gResNet}  and
by using the open-source Tensorflow library \cite{tensorflow2015}. 
The training data sets are usually divided into mini-batches of size $10$. 
All models are trained for
$\sim 300$ epochs with reshuffling after each epoch.
All the weights are initialized randomly from Gaussian
distributions and all the biases are initialized to be zeros. We use
$\sigma(x) = \tanh(x)$ as activation function in all the examples.

\subsection{Linear ODEs}

We first study two linear ODE systems, whose exact solutions are
known.
In both examples, our gResNet networks have 3 hidden layers, each of
which with 30 neurons.

\subsubsection{Example 1}
We consider the following simple linear ODE system:
\begin{equation}
	\label{eq:example1}
	\begin{cases}
		\dot{x}_1=x_1-4x_2,\\
		\dot{x}_2=4x_1-7x_2.
	\end{cases}
\end{equation}
The computational domain is taken to be $D=[0, 2]^2$ and the time lag $\Delta
= 0.1$.  No pre-existing prior model is involved. Instead, we
adopt the approach discussed in Section \ref{sec:DMD} and construct a
standard DMD as the prior model in the gResNet model. Note that the
true dynamical system is homogeneous. Subsequently, the standard DMD
can be highly accurate.

\begin{figure}[htbp]
	\begin{center}
		\includegraphics[width=6cm]{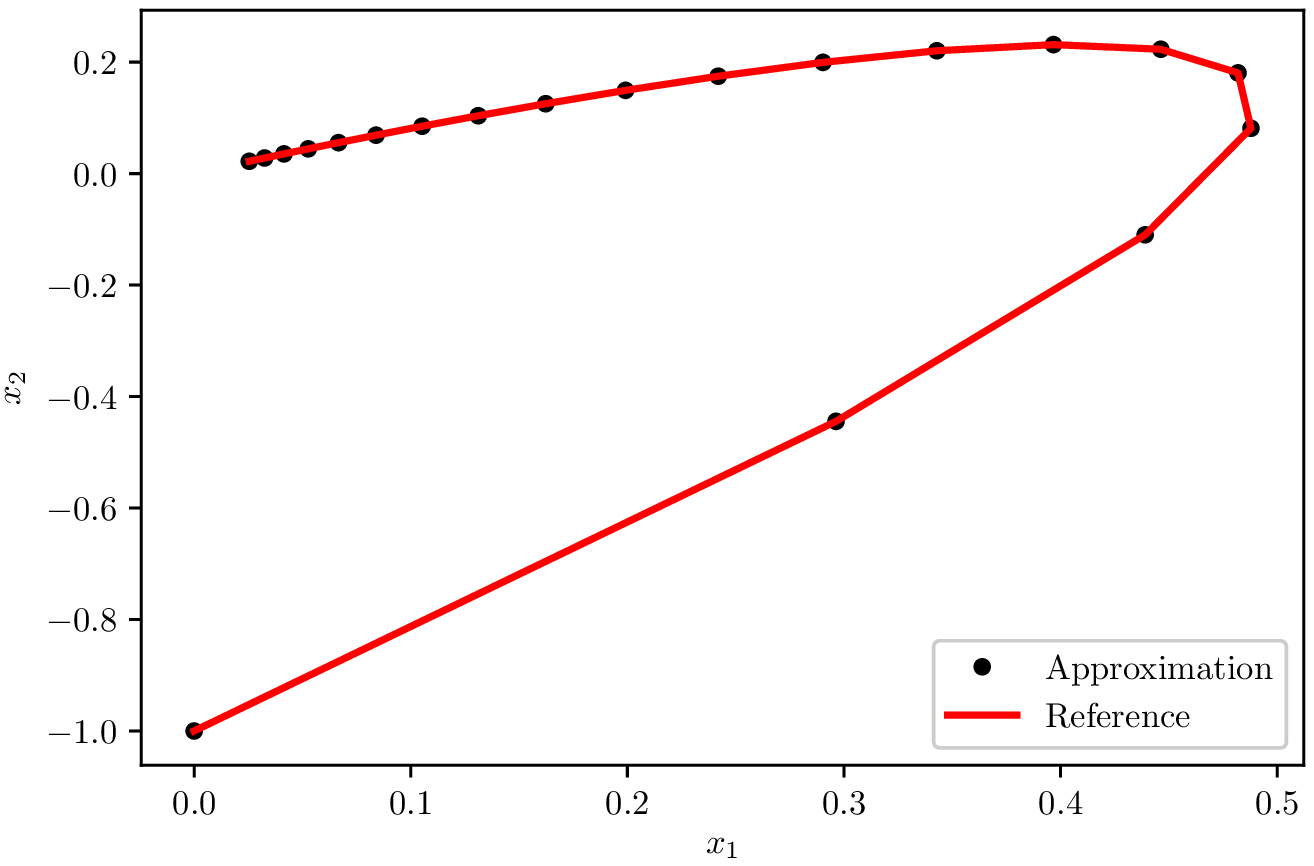}
		\includegraphics[width=6cm]{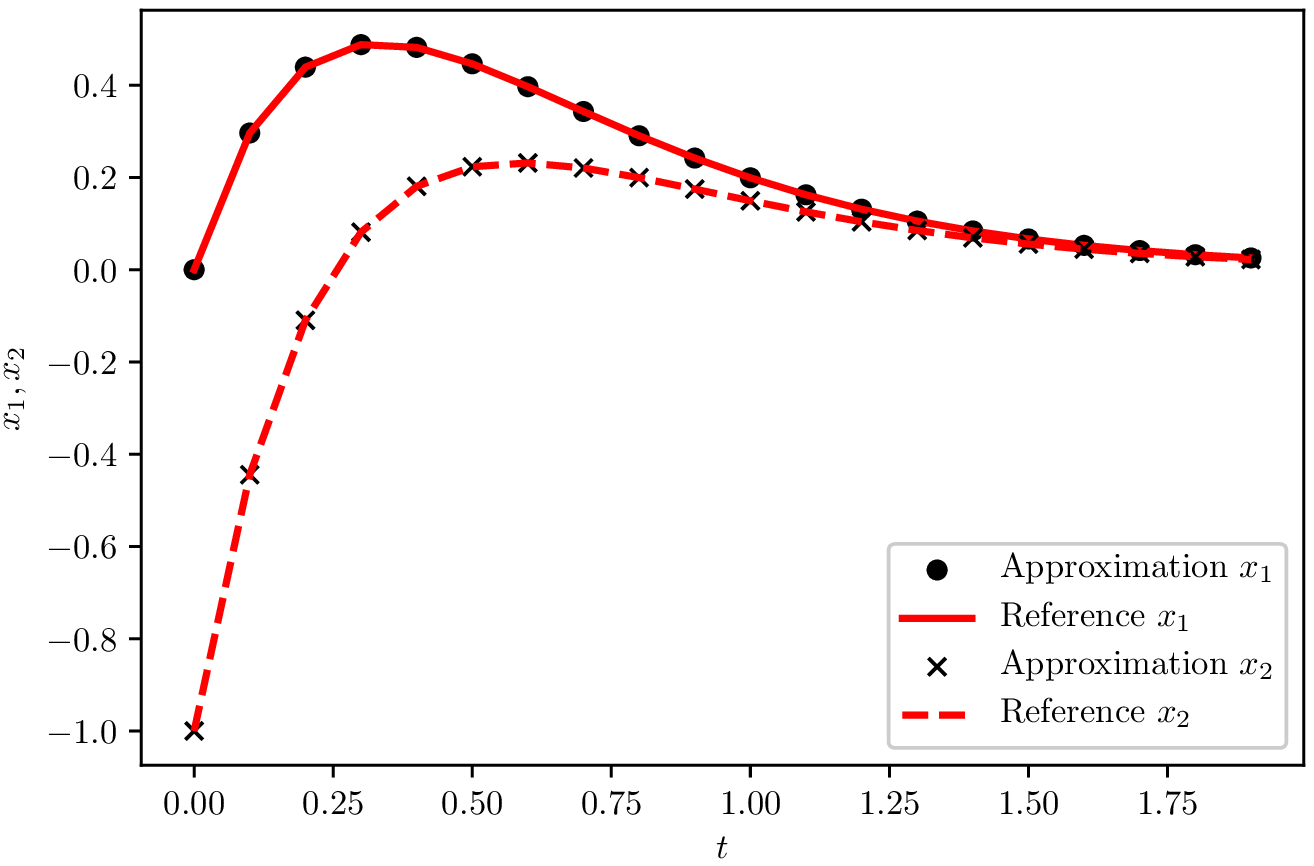}
		\caption{Example 1. Left: phase plot; Right: Trajectory.}
		\label{fig:ex1_traj_train}
	\end{center}
	\begin{center}
		\includegraphics[width=6cm]{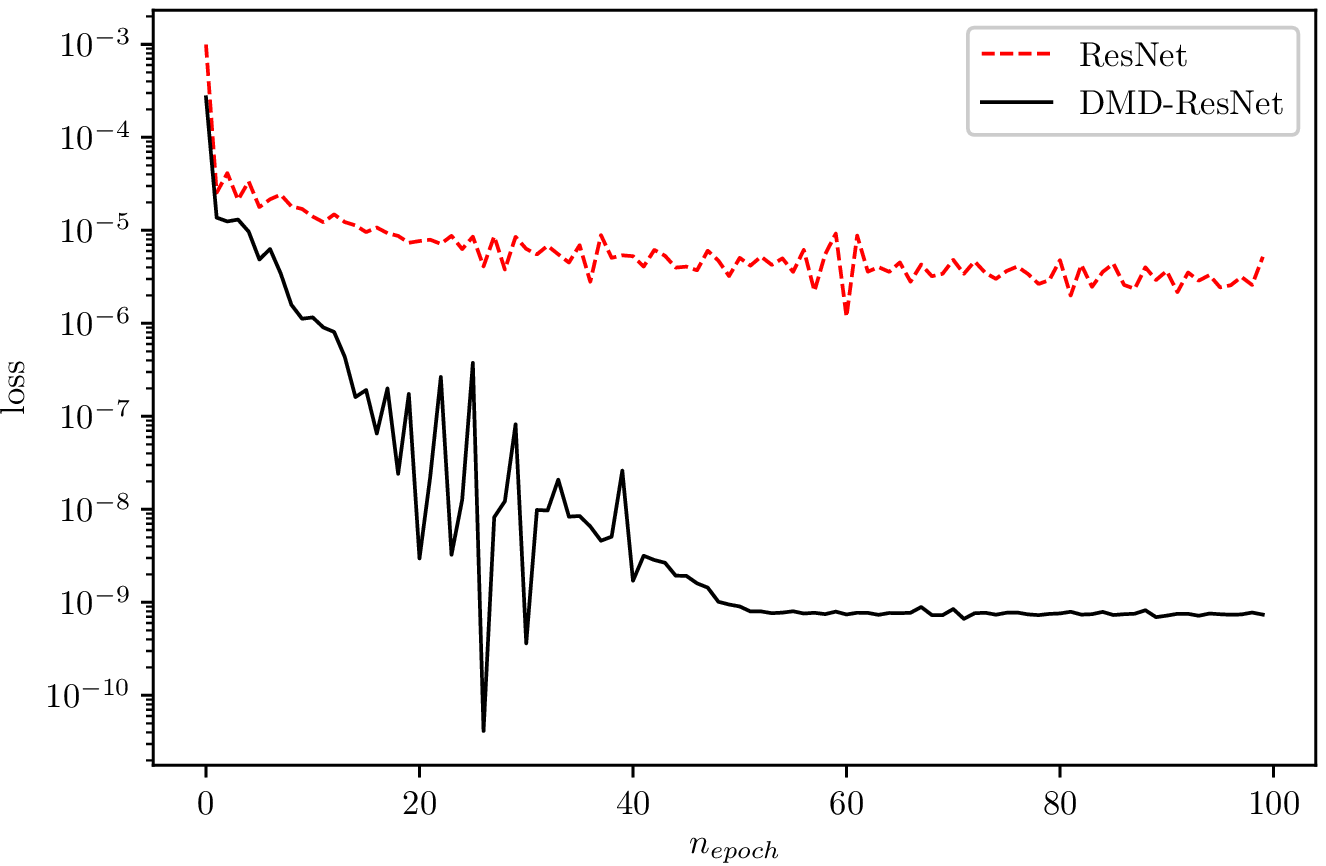}
		\includegraphics[width=6cm]{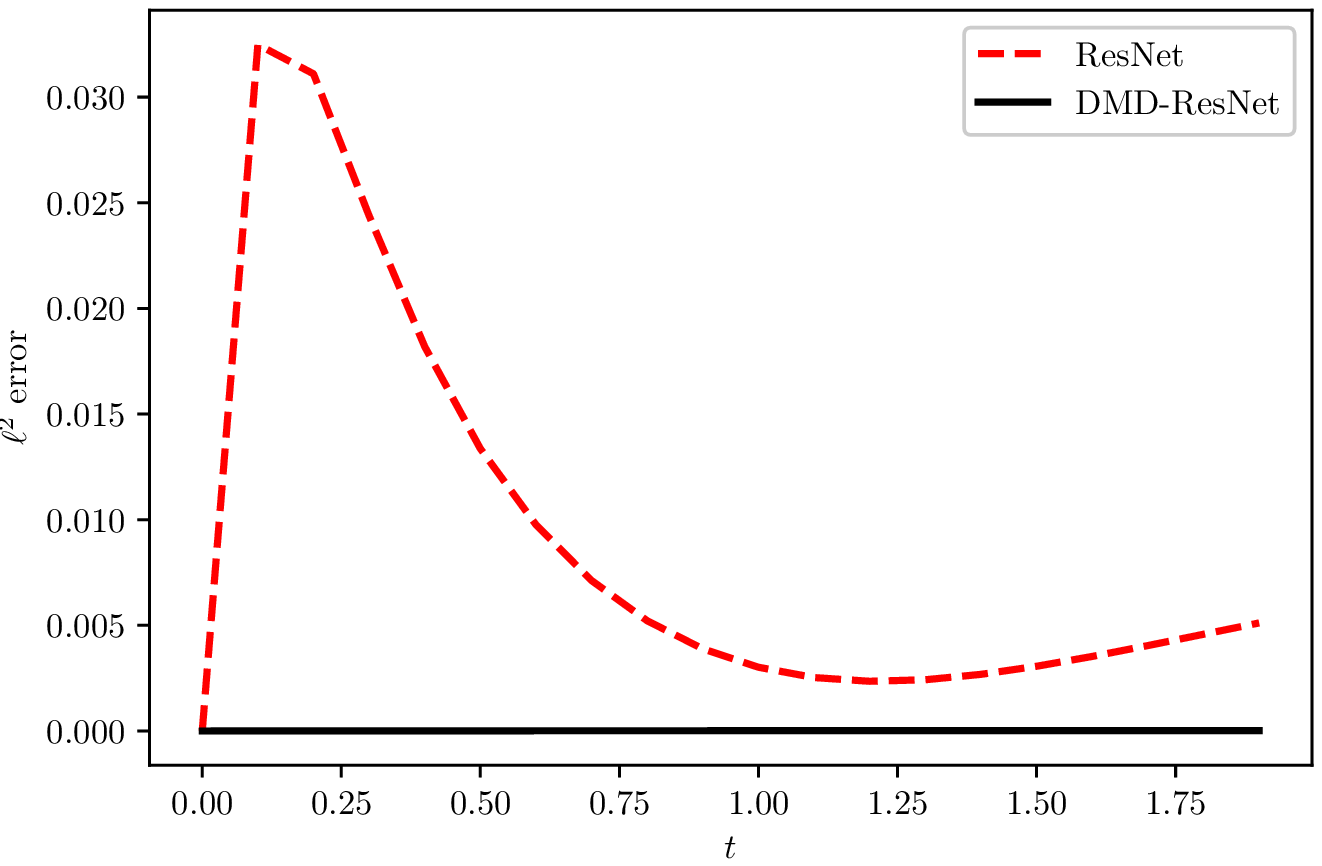}
		\caption{Example 1. Left: Loss history during
                  training; Right: Errors in trajectory prediction.}
		\label{fig:ex1_err_train}
	\end{center}
\end{figure}

After satisfactory training, we conduct system prediction for up to $t
= 2$ for some arbitrarily chosen initial conditions. The phase plot and trajectories of the DMD based gResNet
(denoted as ``DMR-ResNet'') are
shown in \figref{fig:ex1_traj_train}. We observe that the numerical predictions
match the reference solutions extremely well.
This is mostly due to the high accuracy of the DMD prior model. The
neural network, which serves as a correction to the DMD prior model,
has almost negligible impact in this case.
This is manifested from \figref{fig:ex1_err_train} (left),
where we observe the training loss for DMD-ResNet reaches extremely
small magnitude. In the right of \figref{fig:ex1_err_train}, we plot
the numerical errors in the trajectory prediction. We observe that DMD
based gResNet incurs much smaller errors than the standard ResNet.

\subsubsection{Example 2}
\begin{figure}[htbp]
	\begin{center}
		\includegraphics[width=6cm]{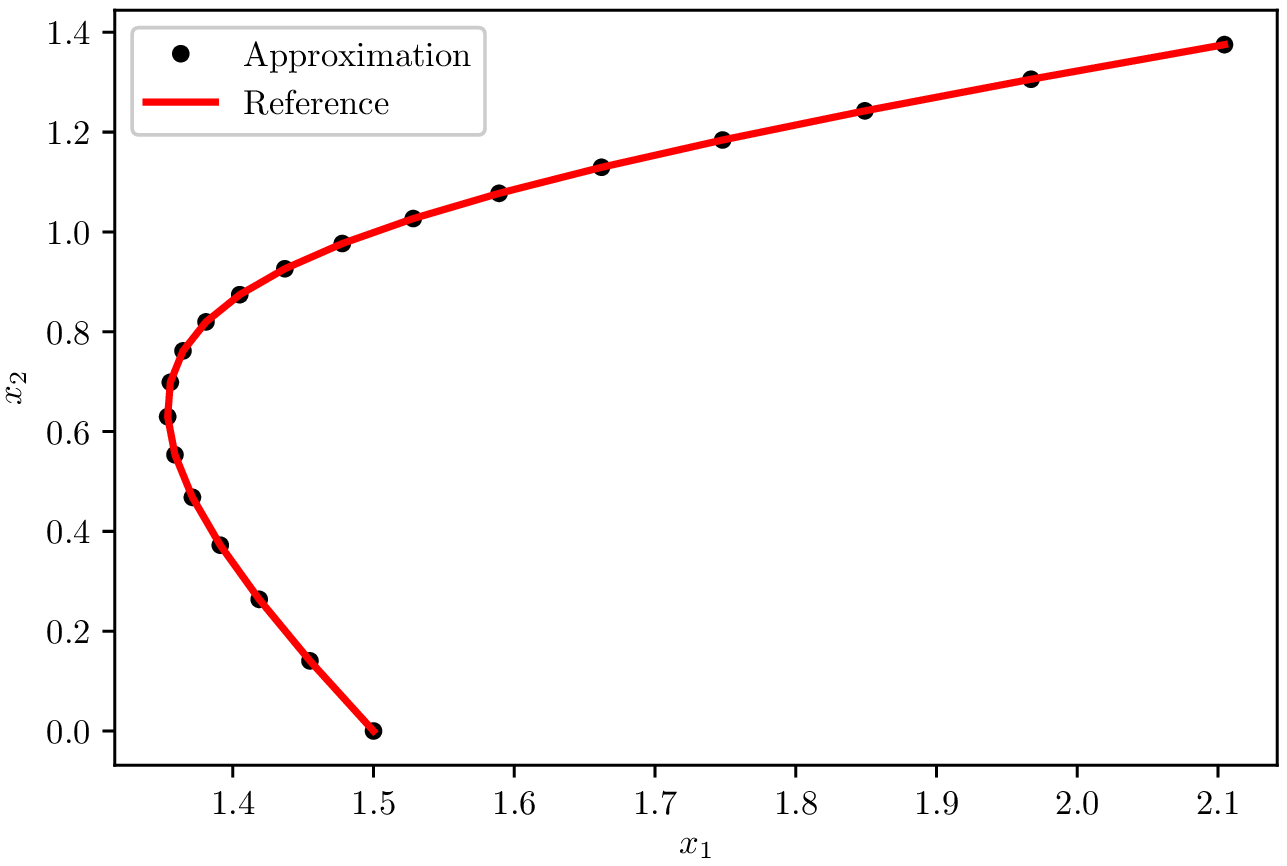}
		\includegraphics[width=6cm]{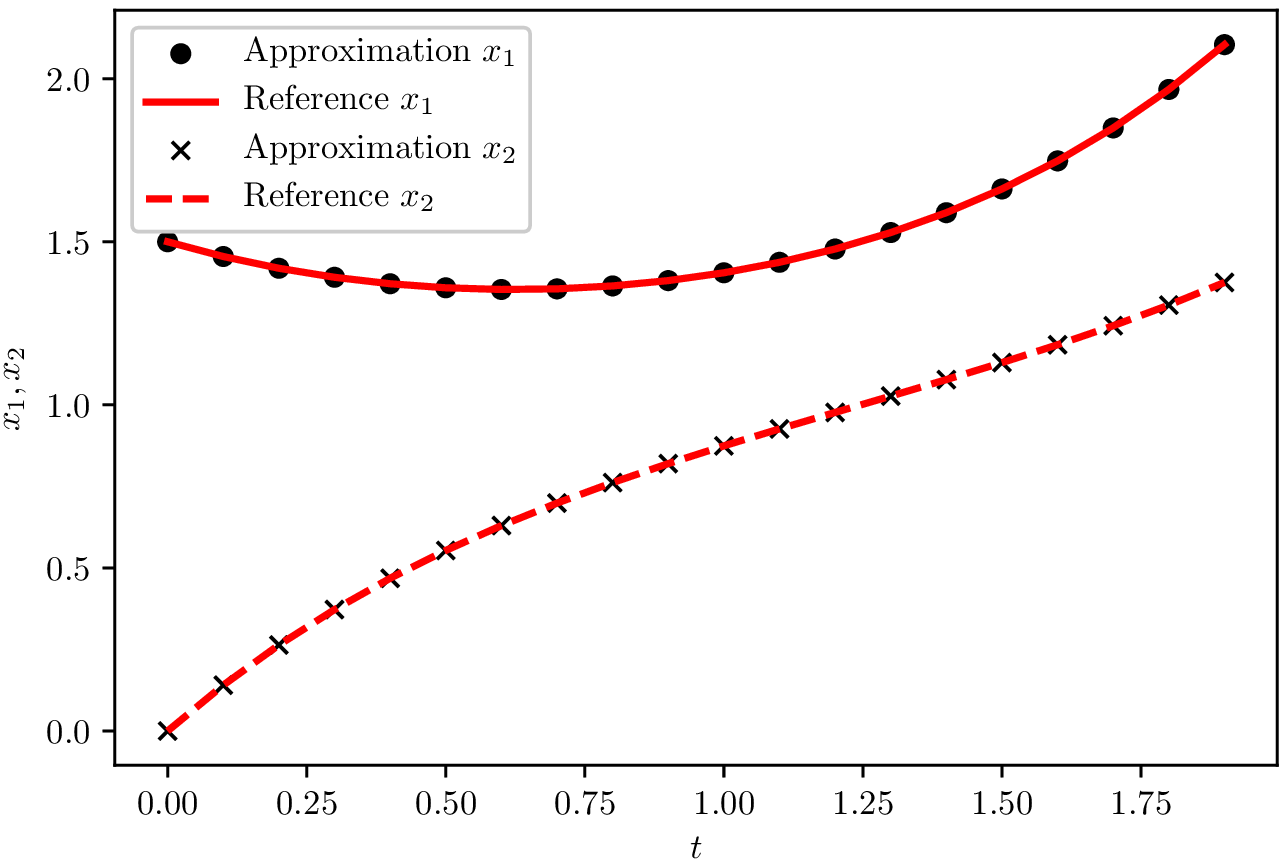}
		\caption{Example 2. Results by mDMD-ResNet.  Left: Phase plot;  Right:
                  Trajectory.}
		\label{fig:ex2_traj_untrain}
	\end{center}
	\begin{center}
		\includegraphics[width=6cm]{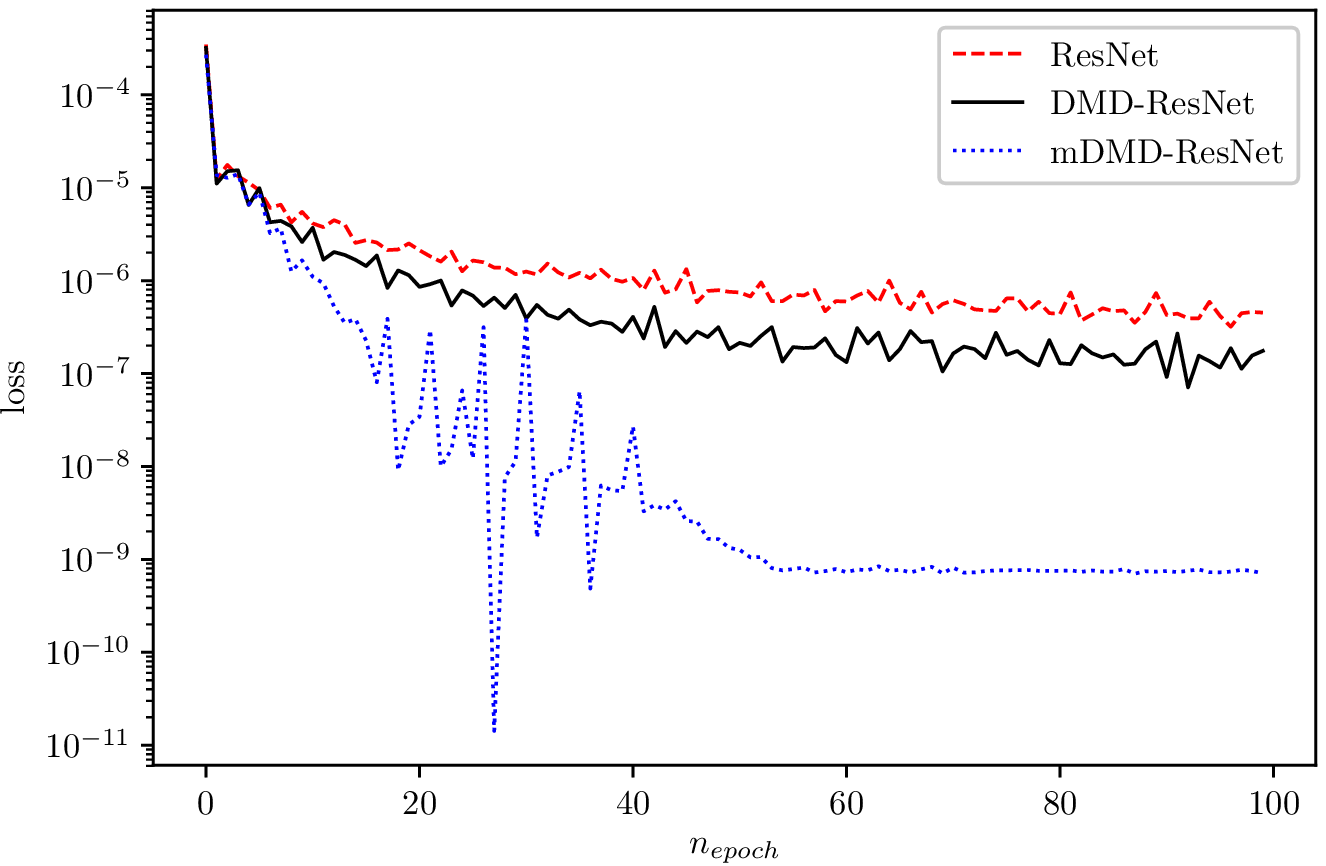}
		\includegraphics[width=6cm]{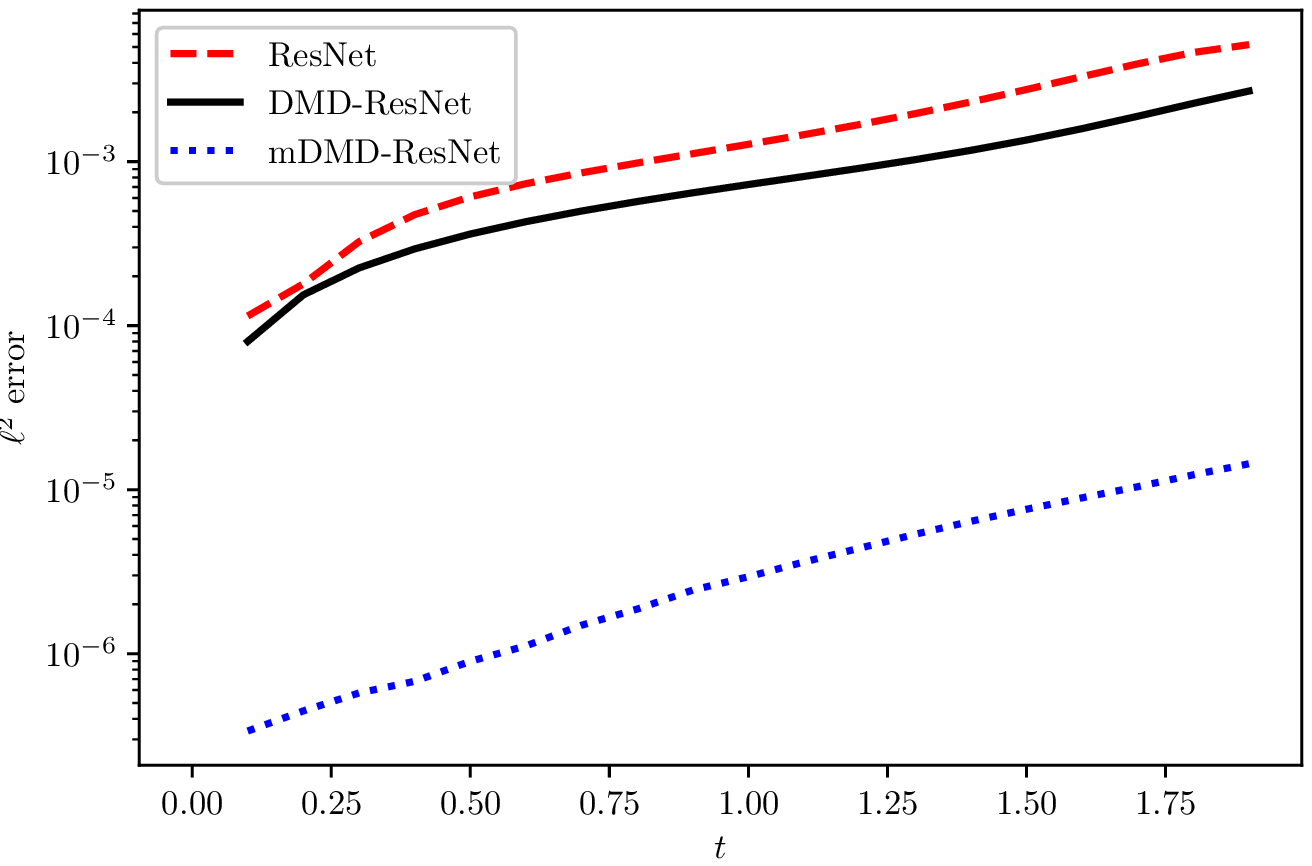}
		\caption{Example 2. Left: Loss history during
                  training; Right: errors in trajectory prediction.}
		\label{fig:ex2_err_untrain}
	\end{center}
\end{figure}

We now consider a non-homogeneous ODE system:
\begin{equation}
\label{eq:example2}
\begin{cases}
\dot{x}_1=x_1+x_2-2,\\
\dot{x}_2=x_1-x_2.
\end{cases}
\end{equation}
The computational domain is taken to be $D=[0, 2]^2$ and the time lag
$\Delta = 0.1$. We employ the standard ResNet, gResNet using the
standard DMD as prior model (DMD-ResNet) and gResNet using the
modified DMD as prior model (mDMD-ResNet). System predictions by the
learned models are conducted for time up to $t=2$. 
In \figref{fig:ex2_traj_untrain}, we show the trajectory plots and
phase portrait produced by the mDMD-ResNet model. We observe very good
agreement with the exact solution. The mDMD-ResNet is in fact the most
accurate model of the three approaches. This can be seen from 
\figref{fig:ex2_err_untrain}. It can be seen that the numerical errors
in the prediction by mDMD-ResNet is two orders of magnitude smaller
than those by ResNet and DMD-ResNet. This demonstrates that the gResNet
method can be advantageous when a proper prior model is available or
can be constructed (in this case via mDMD). The standard DMD is not a
very accurate prior model, as it can not model the non-homogeneous
term in the system. In this case, its performance is similar to the
standard ResNet, which corresponds to using the identity operator as
the prior model.

\begin{figure}[htbp]
	\begin{center}
		\includegraphics[width=6cm]{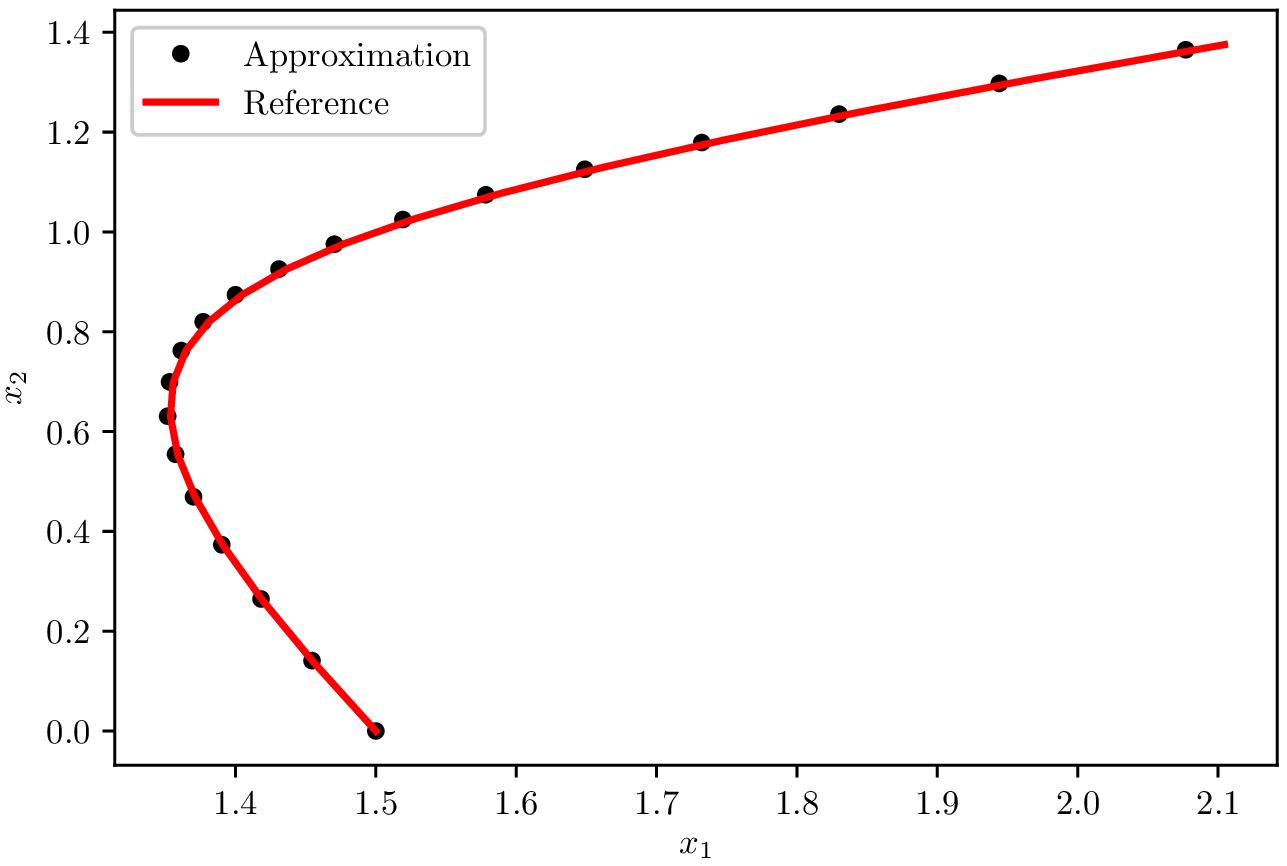}
		\includegraphics[width=6cm]{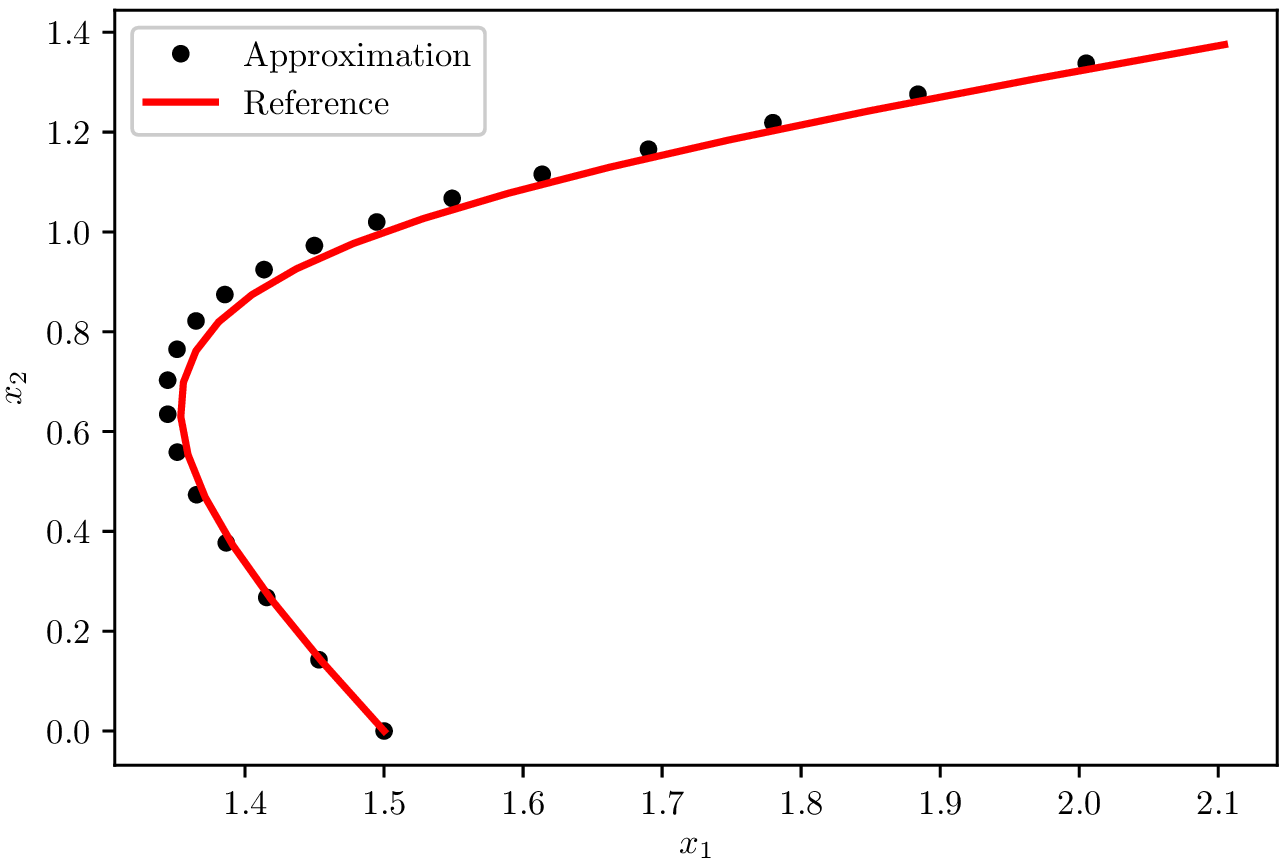}
		\caption{Example 2 with noisy data. Phase plots of mDMD-ResNet with
                  $x_0 = (1.5, 0)$.  Left: $2\%$ noises; Right: $5\%$ noises.}
		\label{fig:ex2_noisy_untrain}
	\end{center}
\end{figure}

We then consider the case of noisy data by adding randomly generated
small noises to the synthetic data.
The results by mDMD-ResNet are shown in
\figref{fig:ex2_noisy_untrain}, with noise at 2\% and 5\%
relative levels. Again, mDMD-ResNet produces accurate system
predictions, whose discrepancy with the exact
is higher at noise level 5\% than at noise level 2\%. This is expected.

\subsection{Nonlinear ODEs}

We now consider four nonlinear examples: (1) a modification of
Example 2 by adding a nonlinear term; (2) the well-known damped
pendulum problem; (3) a nonlinear differential-algebraic equation
(DAE) for electric network model; and (4) the chaotic multiscale
system \eqref{ODE4} from Section \ref{sec:L}.
In the first and fourth example, the neural networks have 3 hidden
layers, each of which with 30 neurons.
In the second and third examples, the networks have 2 hidden layers,
each of which with 40 neurons.

\subsubsection{Example 3}
\begin{figure}[htbp]
	\begin{center}
		\includegraphics[width=6cm]{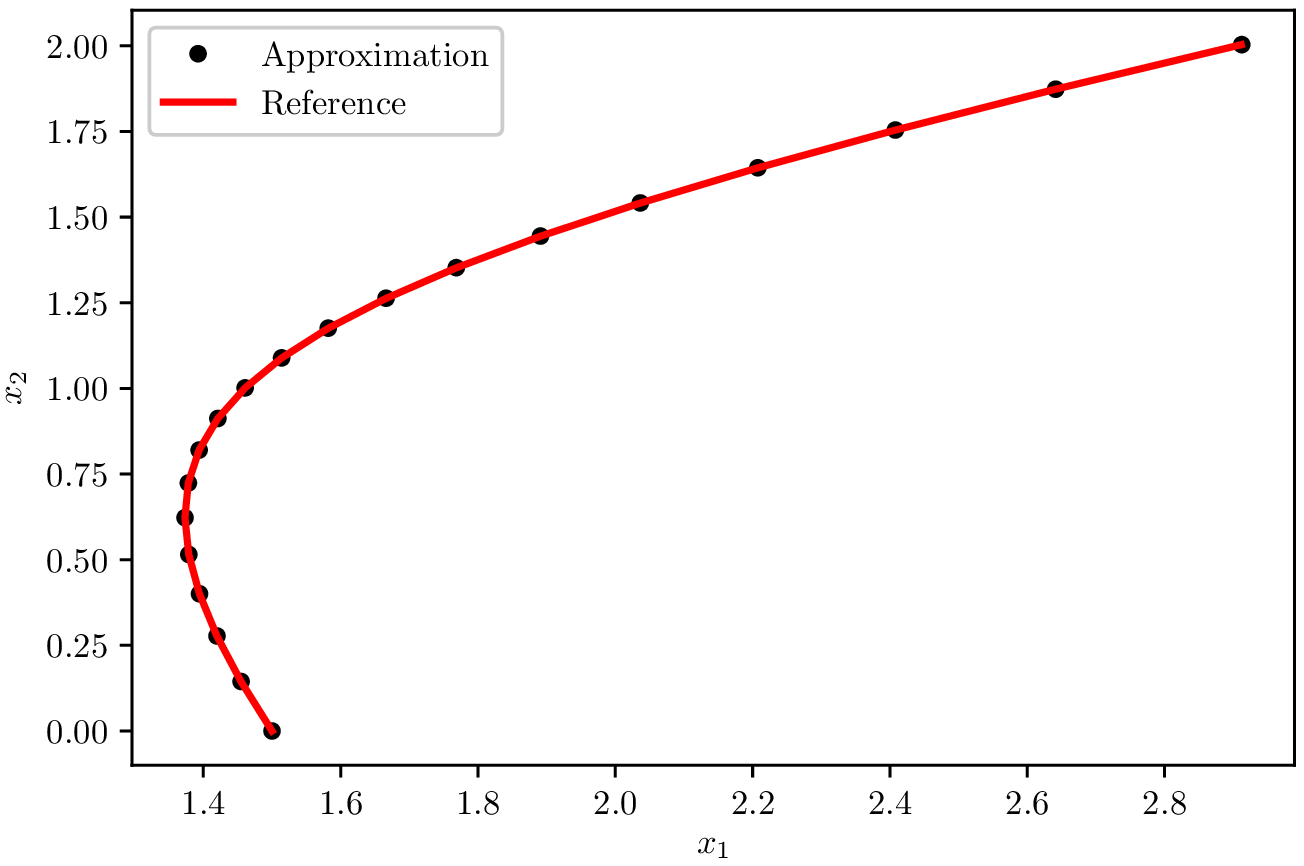}
		\includegraphics[width=6cm]{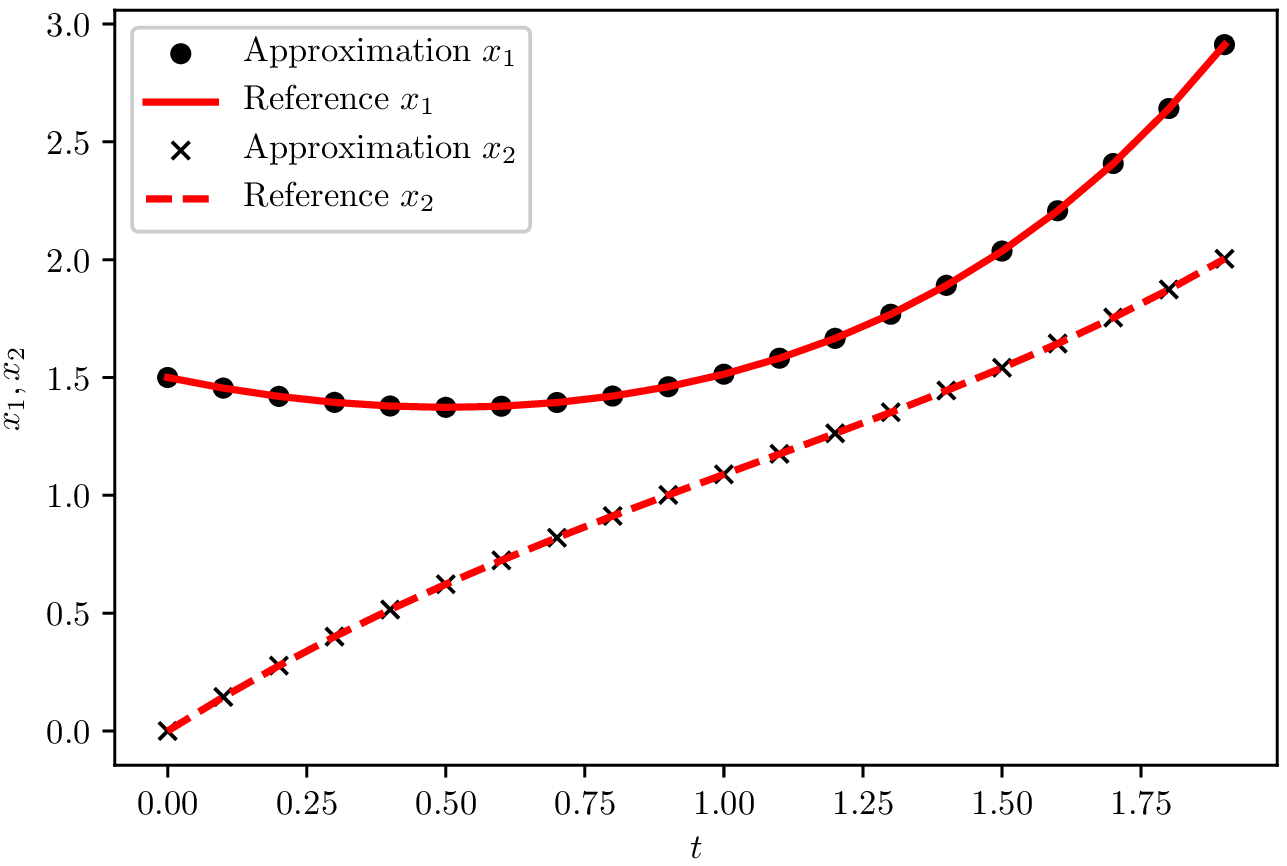}
		\caption{Example 3 with mDMD-ResNet. Left: phase plot
                  with $\x_0 = (1.5, 0)$; Right: Trajectory prediction.}
		\label{fig:ex3_traj_untrain}
	\end{center}
	\begin{center}
		\includegraphics[width=6cm]{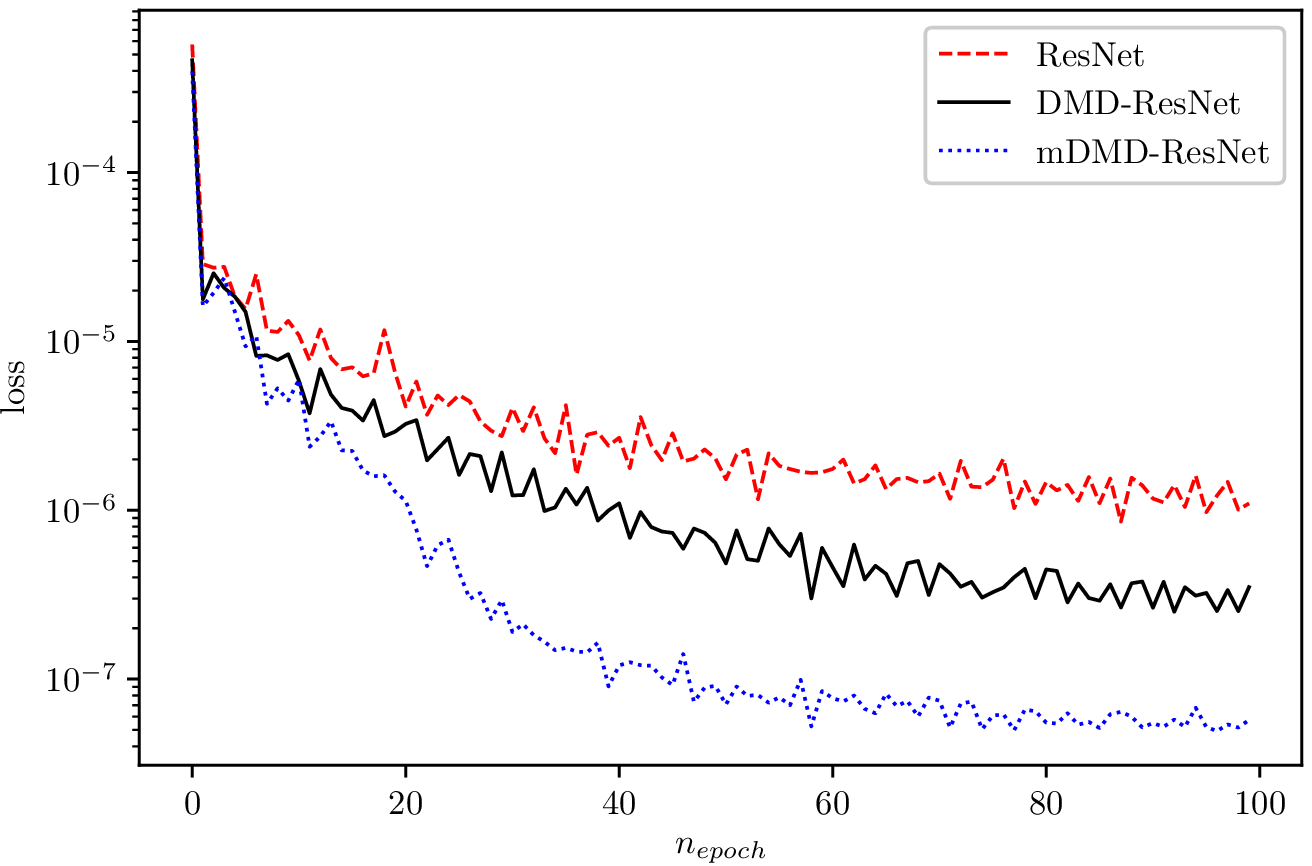}
		\includegraphics[width=6cm]{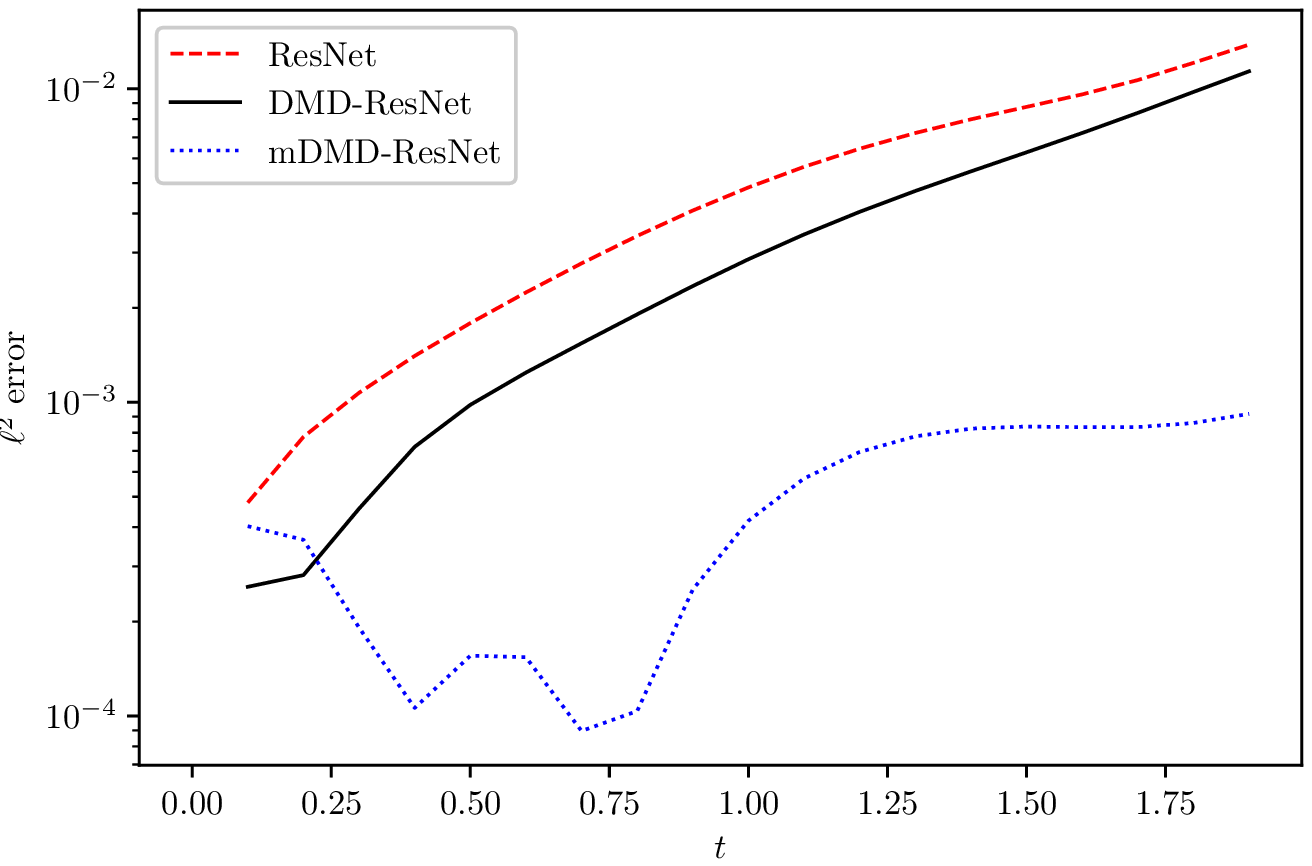}
		\caption{Example 3. Left loss history during training;
                  Right: errors in trajectory prediction.}
		\label{fig:ex3_err_untrain}
	\end{center}
\end{figure}

We consider the following system
\begin{equation}
\label{eq:example3}
\begin{cases}
\dot{x}_1=x_1+x_2-2,\\
\dot{x}_2=x_1-x_2 + 0.5 \sin{x_2}.
\end{cases}
\end{equation}
This is a modification of Example 2 by adding a nonlinear term.
The computational domain is taken to be $D=[0, 3]^2$ and the time lag $\Delta  = 0.1$. 
Upon learning the system, predictions are conducted up to $t=2$.  The
phase plot and trajectories produced by the mDMD-ResNet are plotted in
\figref{fig:ex3_traj_untrain}. The training error history and
numerical errors in the trajectory predictions are plotted In
\figref{fig:ex3_err_untrain}, along with those produced by ResNet and
DMD-ResNet. Again, we observe that mDMD-ResNet produces far superior
predictions than those by ResNet and DMD-ResNet. This is further
demonstrated in Table \ref{table_ex3}. Note that the network norm from
mDMD-ResNet is much smaller than ResNet and DMD-ResNet. This is
primary reason for the better performance by mDMD-ResNet.

\begin{table}[htbp]
	\begin{center}
		\caption{Example 3. Key network properties for ResNet,
                  DMD-ResNet and mDMD-ResNet.}
		\begin{tabular}{ |c|c|c|c|c| }
			\hline
			 & Prediction error & Training loss & Validation loss & Network norm\\  
			 \hline 
			 ResNet & 5.5296$\mathrm{e}-03$ & 7.9243$\mathrm{e}-08$ & 7.8290$\mathrm{e}-08$ & 2.3129$\mathrm{e}-02$ \\
			\hline
			DMD-ResNet & 3.8551$\mathrm{e}-03$  & 1.8546$\mathrm{e}-08$ &  1.8381$\mathrm{e}-08$ &  8.9920$\mathrm{e}-03$\\ 
			\hline
			mDMD-ResNet & 4.9431$\mathrm{e}-04$  & 6.6394$\mathrm{e}-09$ &  6.6046$\mathrm{e}-09$ &  1.4054$\mathrm{e}-03$\\ 
			\hline
		\end{tabular}
		\label{table_ex3}
	\end{center}
\end{table}

\subsubsection{Example 4: Damped pendulum}

\begin{figure}[htbp]
	\begin{center}
		\includegraphics[width=6cm]{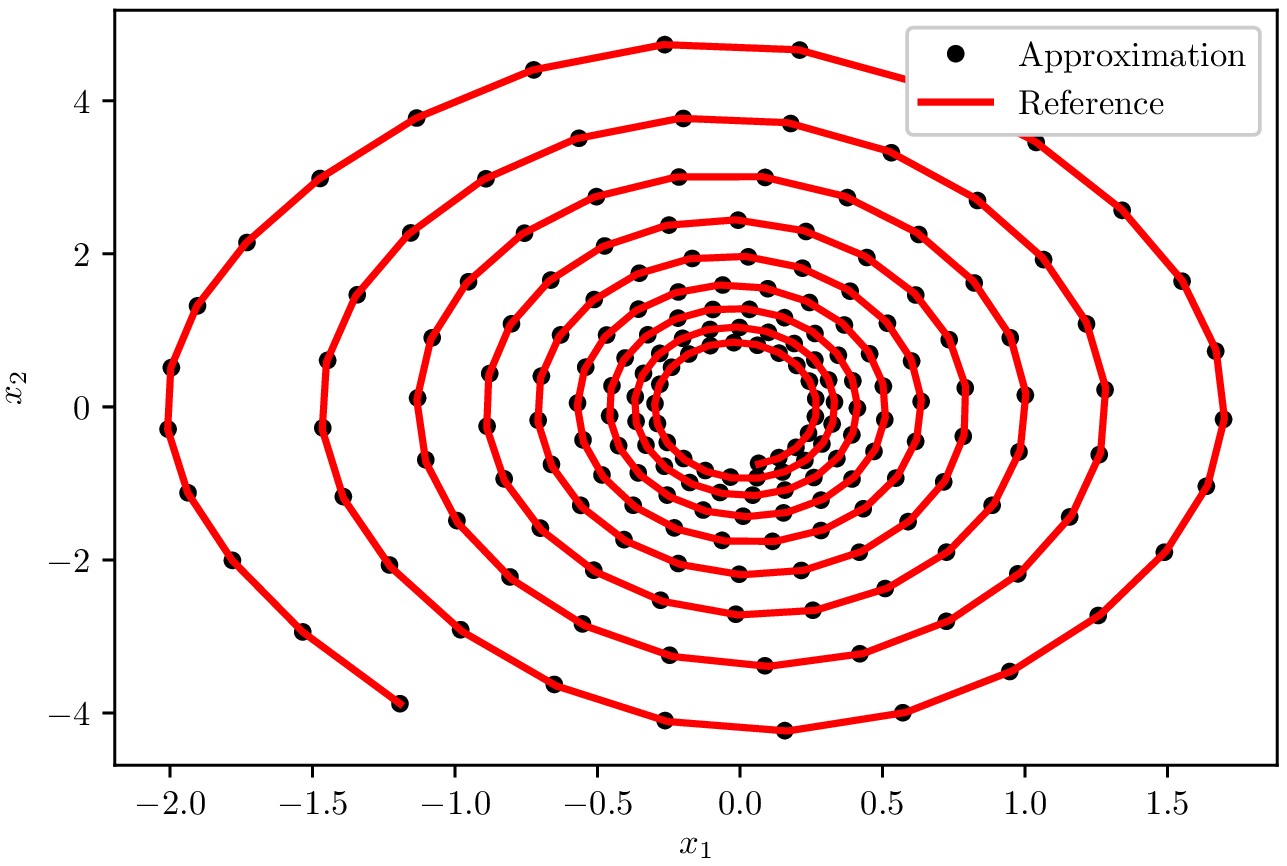}
		\includegraphics[width=6cm]{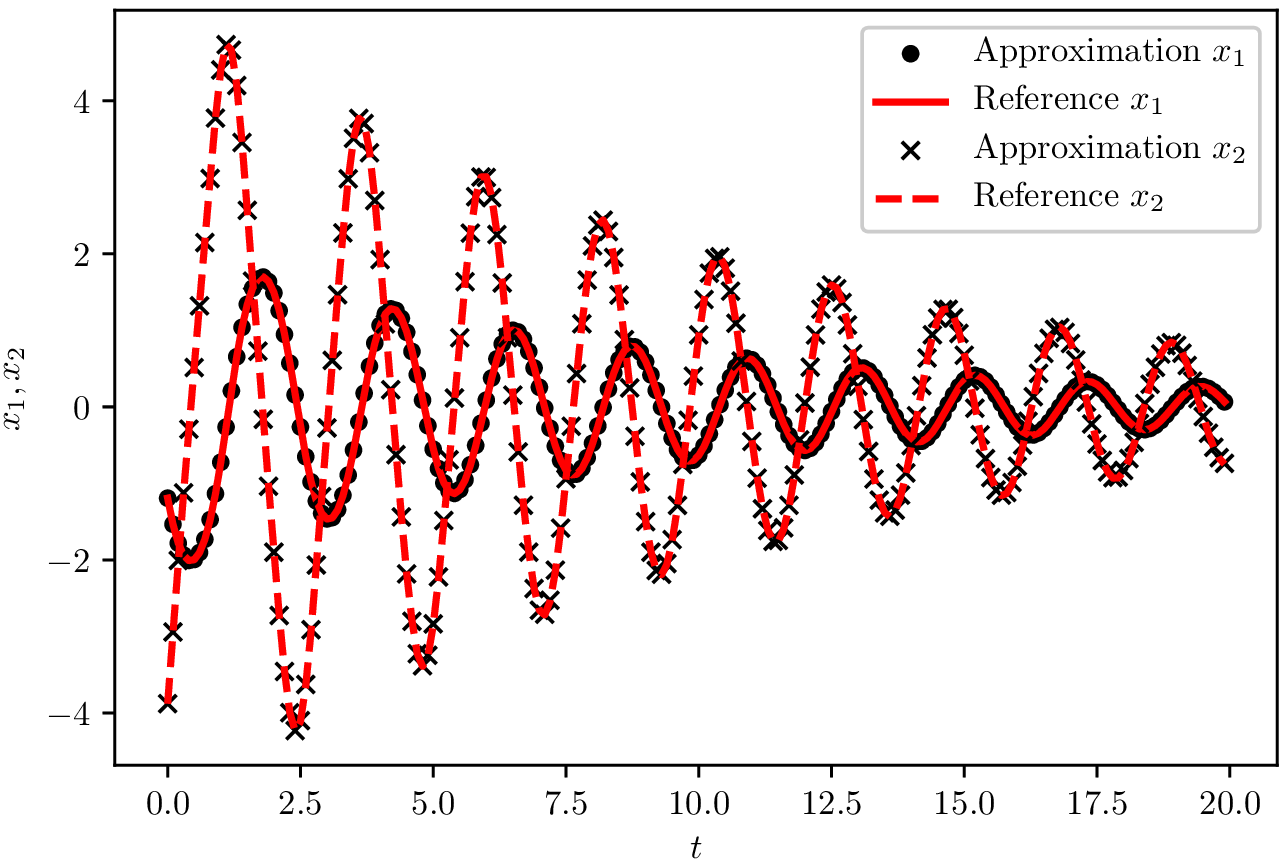}
		\caption{Example 4 with adaptive gResNet
                  modeling. Left: phase portrait; Right: trajectory prediction.}
		\label{fig:ex4_traj_train}
	\end{center}
	\begin{center}
		\includegraphics[width=6cm]{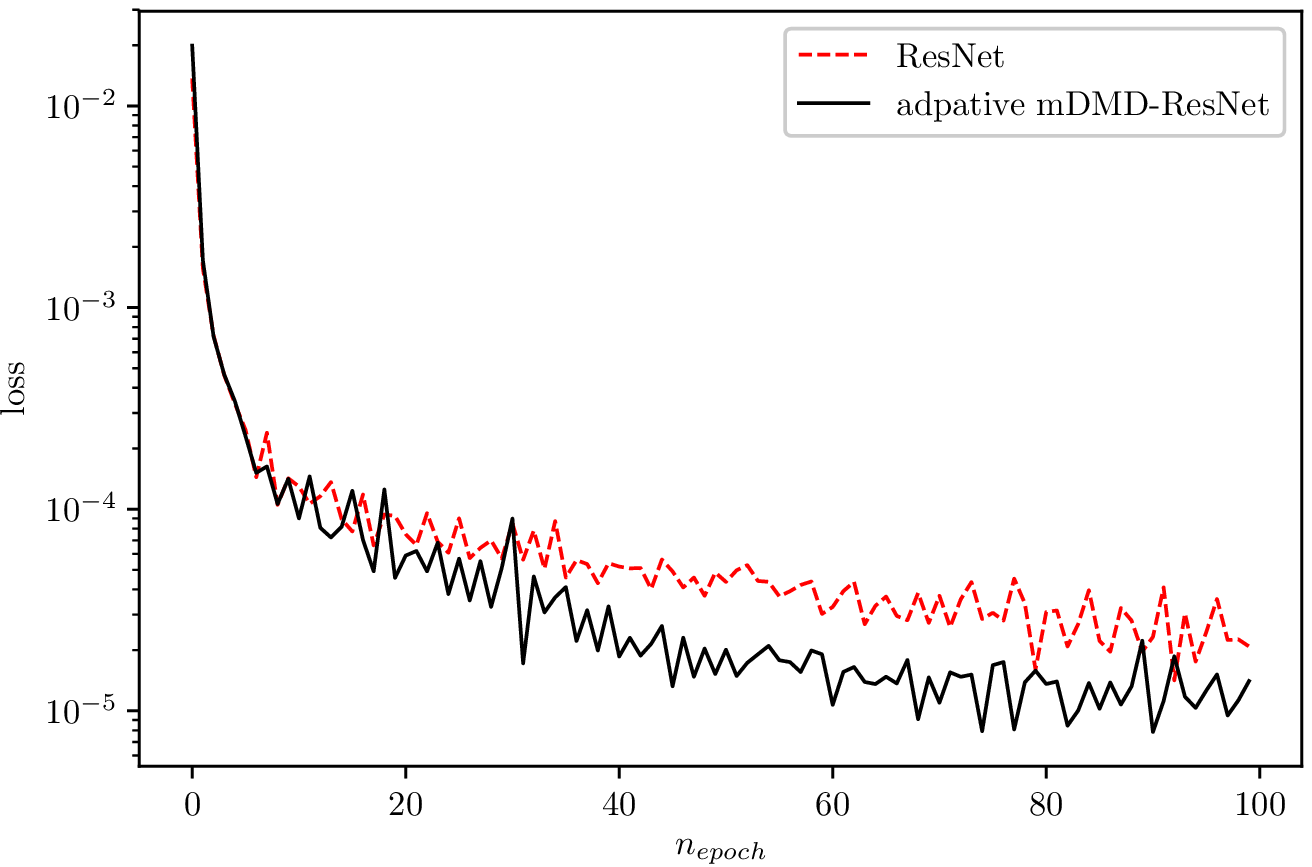}
		\includegraphics[width=6cm]{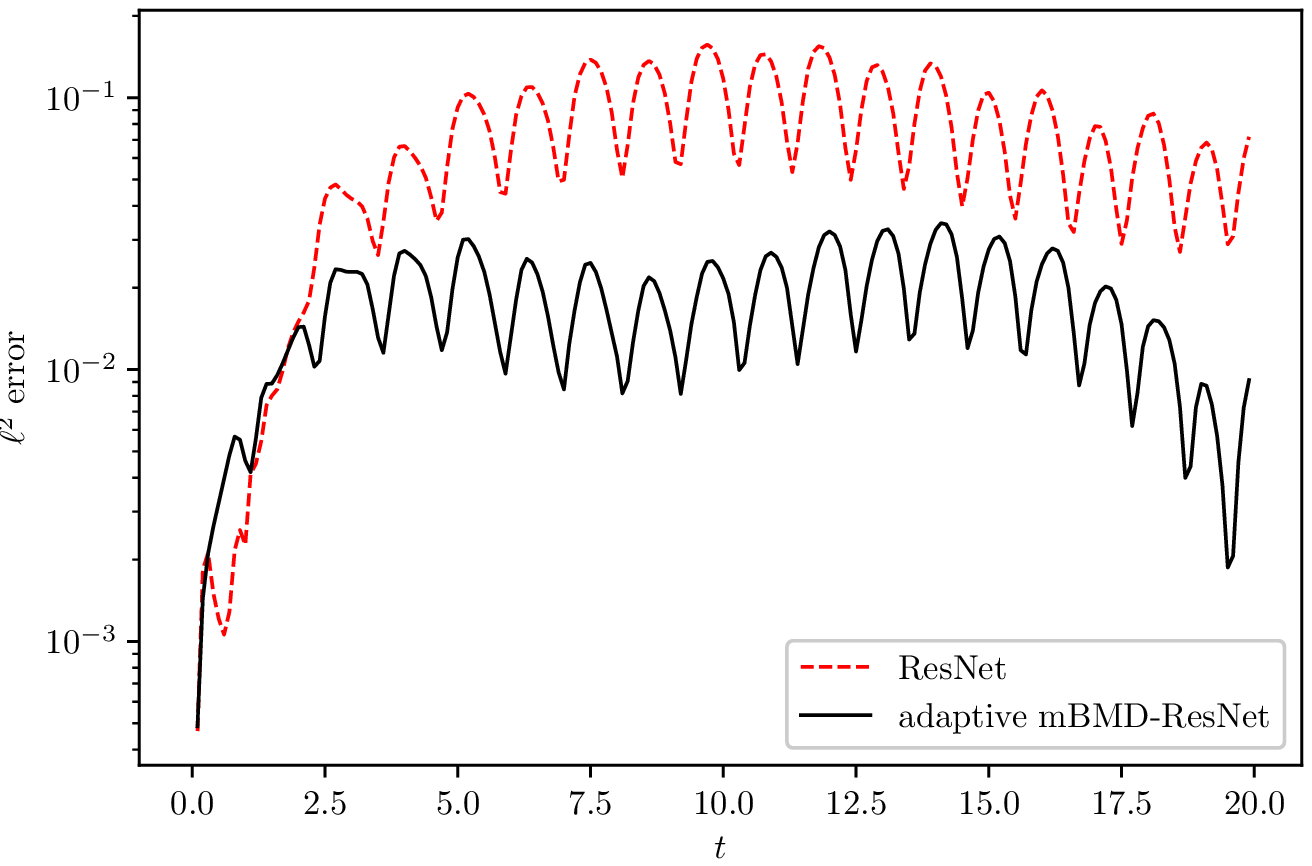}
		\caption{Example 4. Left Loss history during training;
                  Right: errors in trajectory prediction.}
		\label{fig:ex4_err}
	\end{center}
\end{figure}

We now consider the damped pendulum problem
\begin{equation*}
	\begin{cases}
		\dot{x}_1=x_2,\\
		\dot{x}_2=-\alpha x_2-\beta \sin x_1,
	\end{cases}
\end{equation*}
where $\alpha=0.2$ and $\beta=8.91$. The computational domain is $D=[-\pi,
\pi]\times [-2\pi, 2\pi]$, and the time lag $\Delta = 0.1$. Here we
employ the adaptive nonlinear approximation method from Section
\ref{sec:adaptiveL} as the prior model.
In \figref{fig:ex4_traj_train}, we present the trajectory prediction results of the
adaptive gResNet model, starting from an arbitrarily chosen initial condition $\x_0=(-1.193, -3.876)$ and
for time up to $t=20$. We observe excellent agreements between the network model and the
reference solution. In \figref{fig:ex4_err}, we show training loss
history and trajectory errors in the prediction, along with those
obtained by ResNet. It can be seen that, with errors in
prediction one order of magnitude smaller, the adaptive gResNet produces
much more accurate results than the standard ResNet.


\subsubsection*{Example 5: Nonlinear electric network}
We now consider a system of nonlinear differential-algebraic
equations (DAE), a  model for a electric network from \cite{pulch2013polynomial},
\begin{equation*}
	\begin{cases}
		\dot{u}_1 = v_2/C, \\
		\dot{u}_2 = u_1/L, \\
		0 = v_1 - (G_0 - G_{\infty})U_0 \tanh(u_1/U_0) - G_{\infty}u_1, \\
		0 = v_2 + u_2 + v_1, \\
	\end{cases}
\end{equation*}
where $u_1$ denotes the node voltage, and $u_2, v_1$ are branch
currents. The physical parameters are specified as $C=10^{-9}, L =
10^{-6}, U_0 = 1, G_0 = -0.1$ and $G_{\infty} = 0.25$. In our test, we
define the computational domain of $(u_1, u_2)$ as $D = [-2,2] \times
[-0.2, 0.2]$ and fix the time lag  $\Delta t = 2 \times 10^{-9}$. For
system prediction, we choose an (arbitrary) initial condition
$\mathbf{u}_0 = (0, 0.1)$ and produce result for  up to $t = 1 \times
10^{-6}$ (1,000 times of the size of $\Delta$).

\begin{figure}[htbp]
	\begin{center}
		\includegraphics[width=6cm]{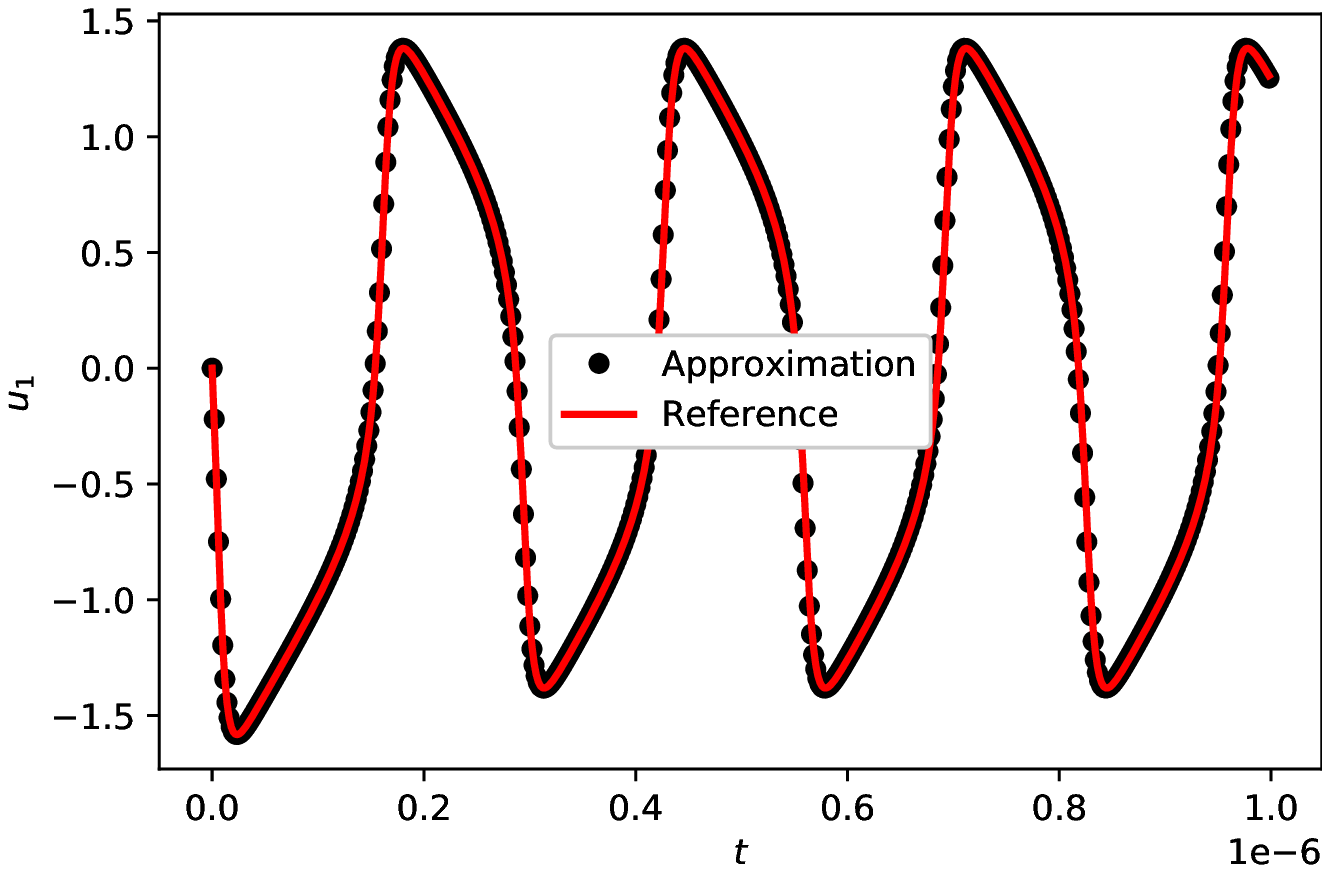}
		\includegraphics[width=6cm]{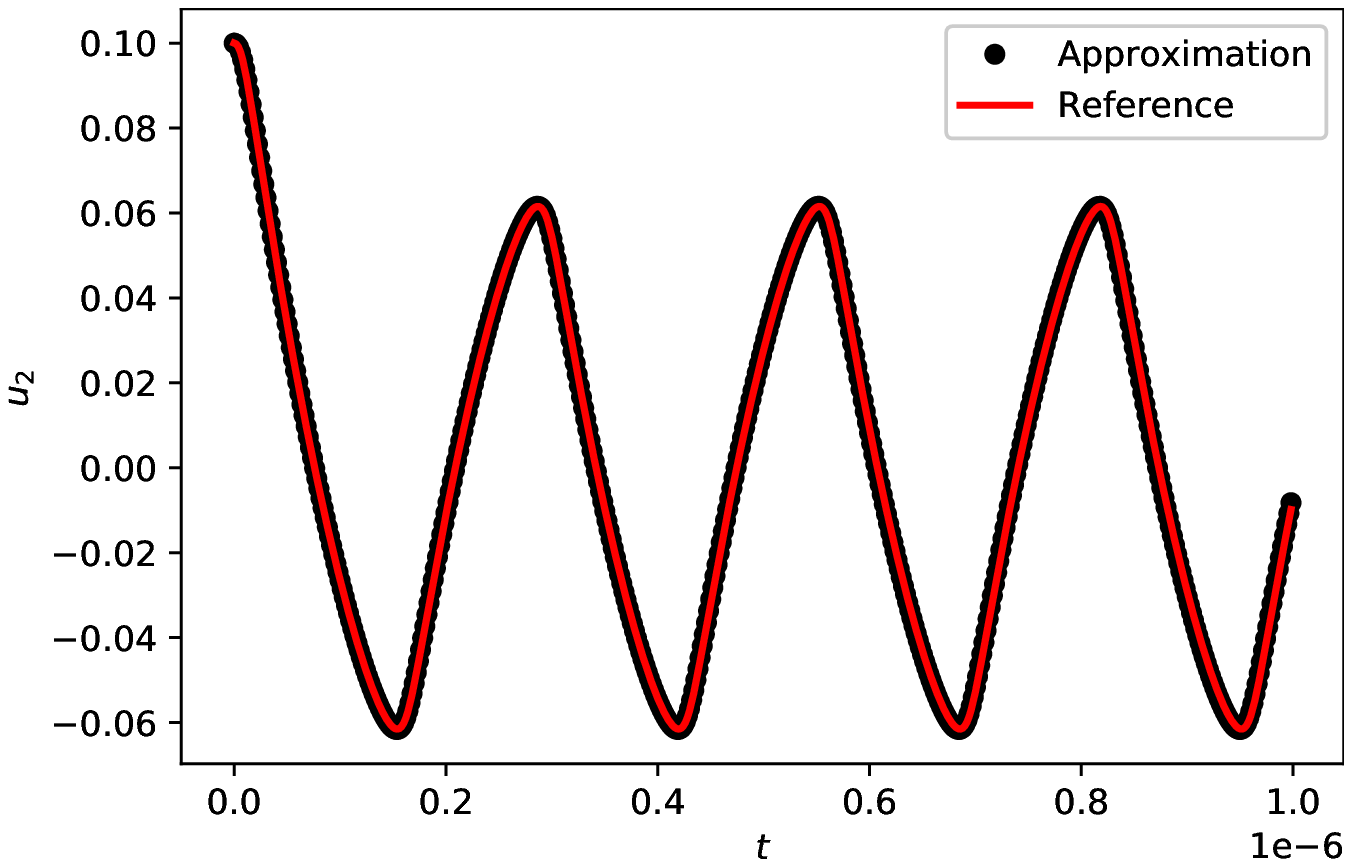}
		\caption{Example 5. Solution prediction of  $(u_1, u_2)$ via mDMD-ResNet.}
		\label{fig:ex5_traj_x1x2_train1}
	\end{center}
	\begin{center}
		\includegraphics[width=6cm]{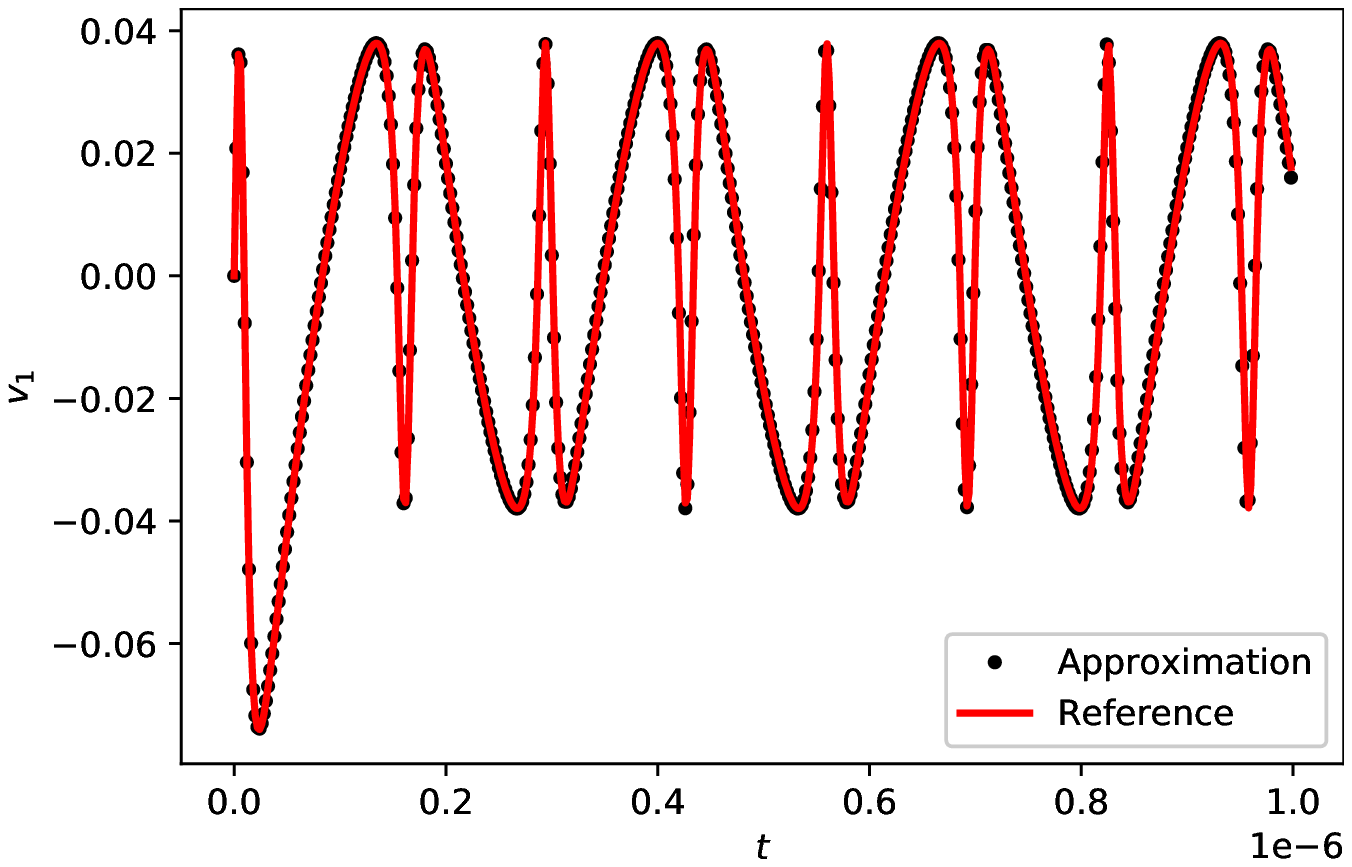}
		\includegraphics[width=6cm]{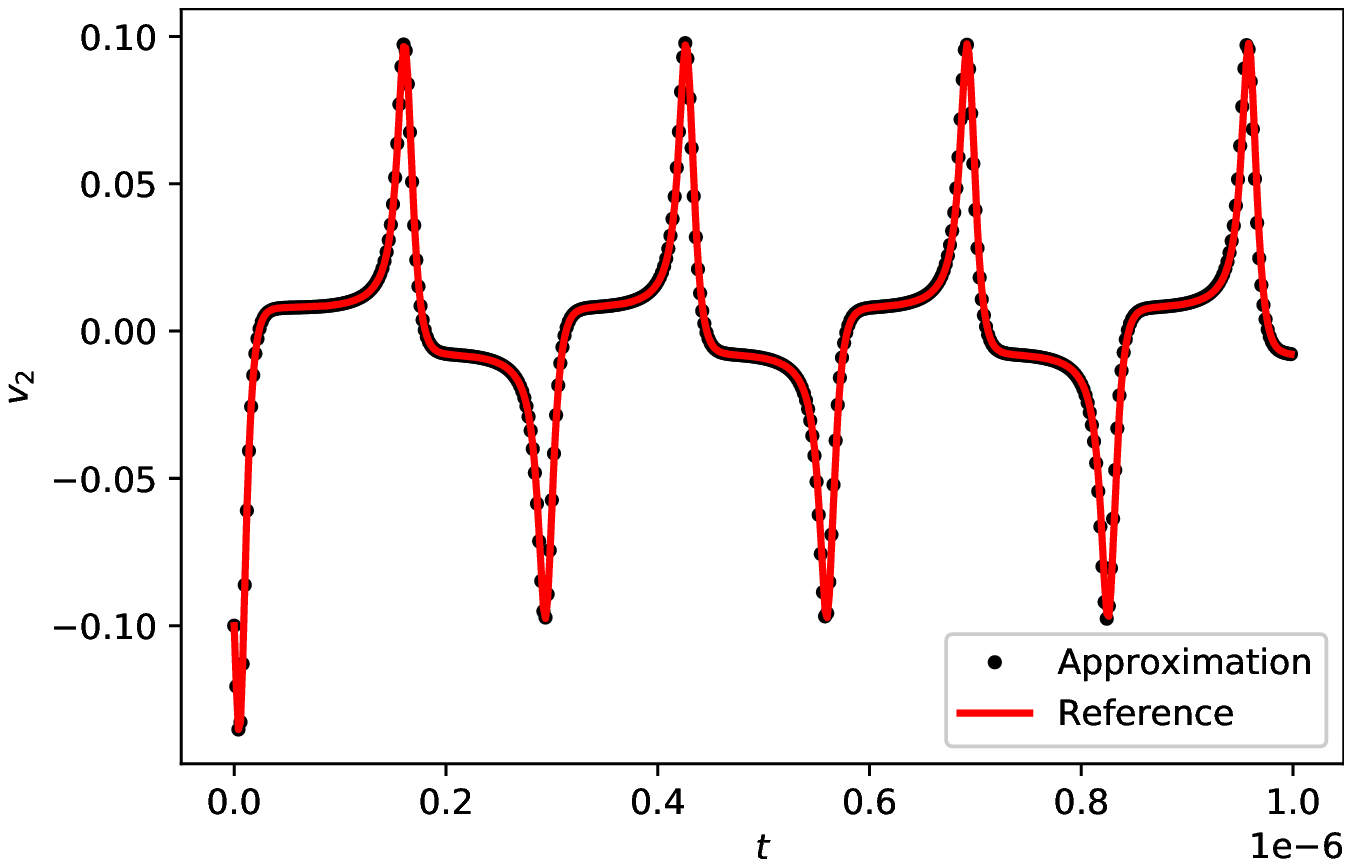}
		\caption{Example 5. Solution prediction of $(v_1,
                  v_2)$ via mDMD-ResNet.}
		\label{fig:ex5_traj_v1v2_train1}
	\end{center}
\end{figure}

\begin{figure}[htbp]
	\begin{center}
		\includegraphics[width=6cm]{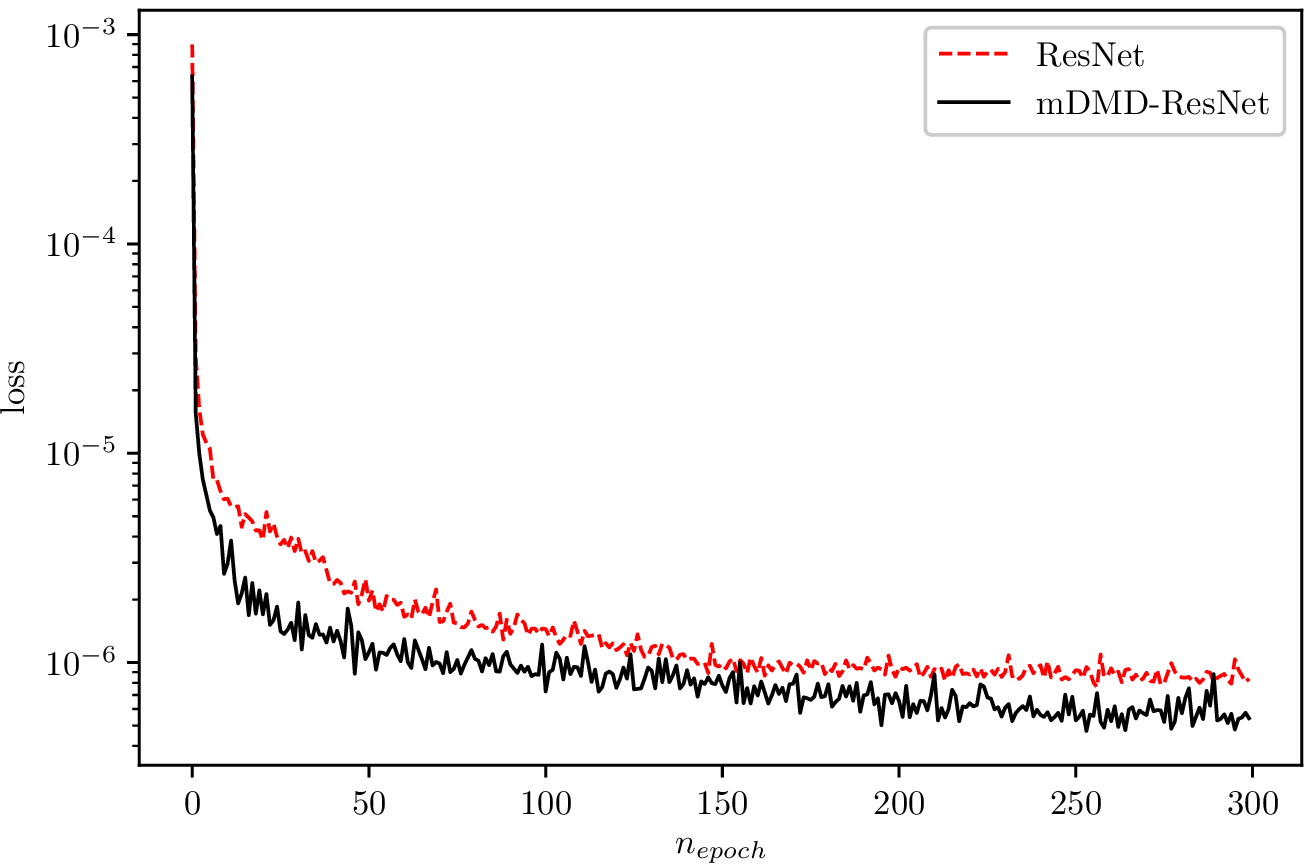}
		\includegraphics[width=6cm]{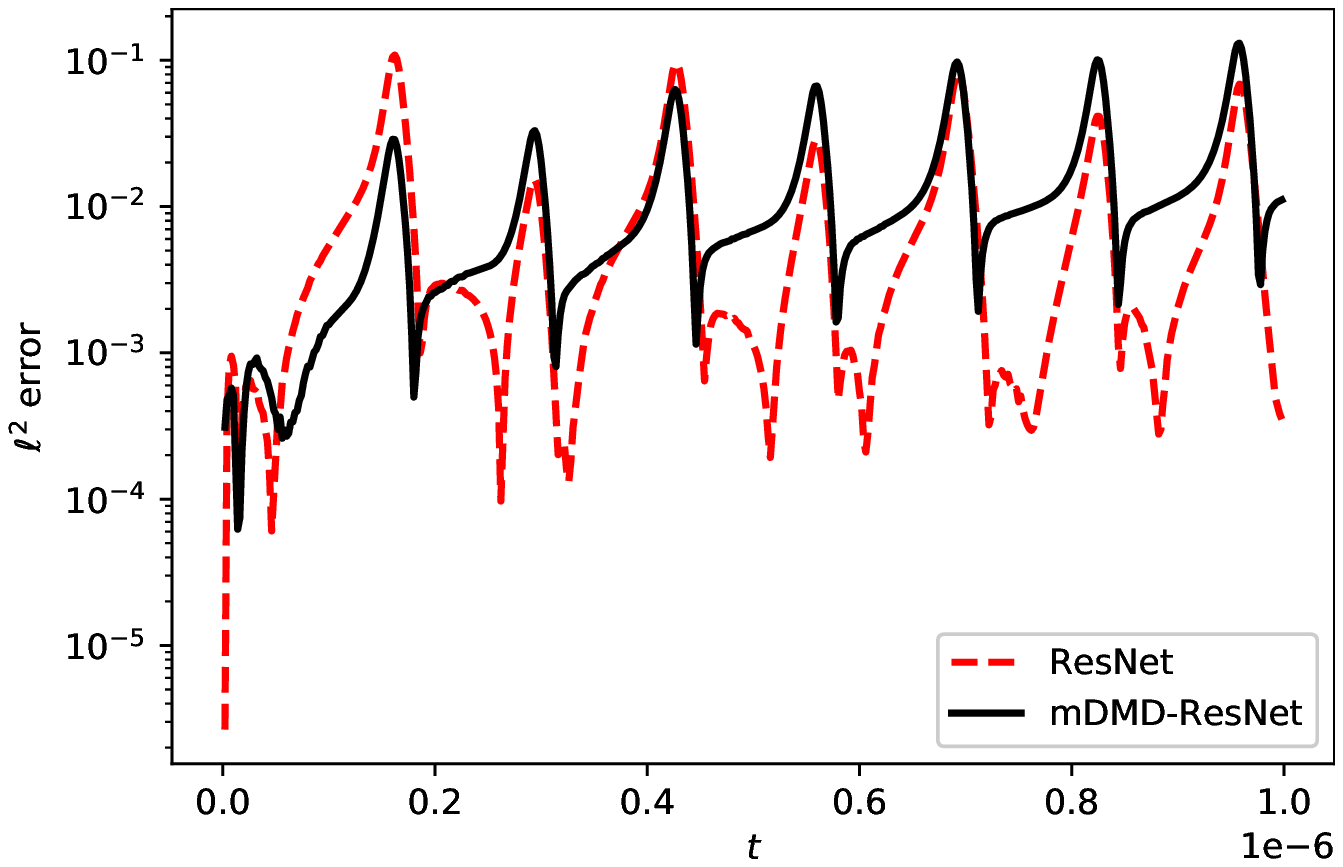}
		\caption{Example 5. Left: loss history during
                  training; Right: errors in trajectory predictions.}
		\label{fig:ex5_error_train}
	\end{center}
\end{figure}

The solution trajectories produced by mDMD-ResNet are shown in
\figref{fig:ex5_traj_x1x2_train1} and
\figref{fig:ex5_traj_v1v2_train1}. We observe high accuracy in the
prediction, when compared with the reference solution. The training
loss history and numerical errors in the system prediction are plotted
in \figref{fig:ex5_error_train}, along with those produced by the
standard ResNet. It can be seen that the performance of ResNet and
DMD-ResNet is similar.
This is because in this particular example the time lag
$\Delta=10^{-9}$ is very small, which is dictated by the scaling of the
physical problem. Consequently, the identity operator $\mathcal{I}$ can be considered a very good
prior model and the standard ResNet performs well. A more 
detailed comparison of the ResNet and mDMD-ResNet models are presented in Table
\ref{table_ex5}. It can be seen that the mDMD-ResNet still offers
slight advantage.


\begin{table}[htbp]
	\begin{center}
		\caption{Example 5: Key network properties for ResNet
                  and mDMD-ResNet.}
		\begin{tabular}{ |c|c|c|c|c| }
			\hline
			& Prediction error & Training loss & Validation loss & Network norm\\  
			\hline 
			ResNet & 1.0724$\mathrm{e}-02$ & 9.67135$\mathrm{e}-08$ & 1.0063$\mathrm{e}-07$ & 2.4569$\mathrm{e}-01$ \\
			\hline
			mDMD-ResNet & 1.4715$\mathrm{e}-02$  & 2.9232$\mathrm{e}-08$ &  3.0879$\mathrm{e}-08$ &  8.3937$\mathrm{e}-02$\\ 
			\hline
		\end{tabular}
		\label{table_ex5}
	\end{center}
\end{table}

\subsubsection*{Example 6}

We now consider the chaotic multiscale
system \eqref{ODE4} from Section \ref{sec:L}. The prior model is the
averaged system \eqref{ODE3}. This represents a case discussed in
Section \ref{sec:L}, where the prior
model is available as an existing coarse model. The operator
$\mathcal{L}$ associated with the prior model
does not have an explicit form and needs to be computed via
numerically solving the reduced system \eqref{ODE3}.

\begin{figure}[htbp]
	\begin{center}
		\includegraphics[width=12cm]{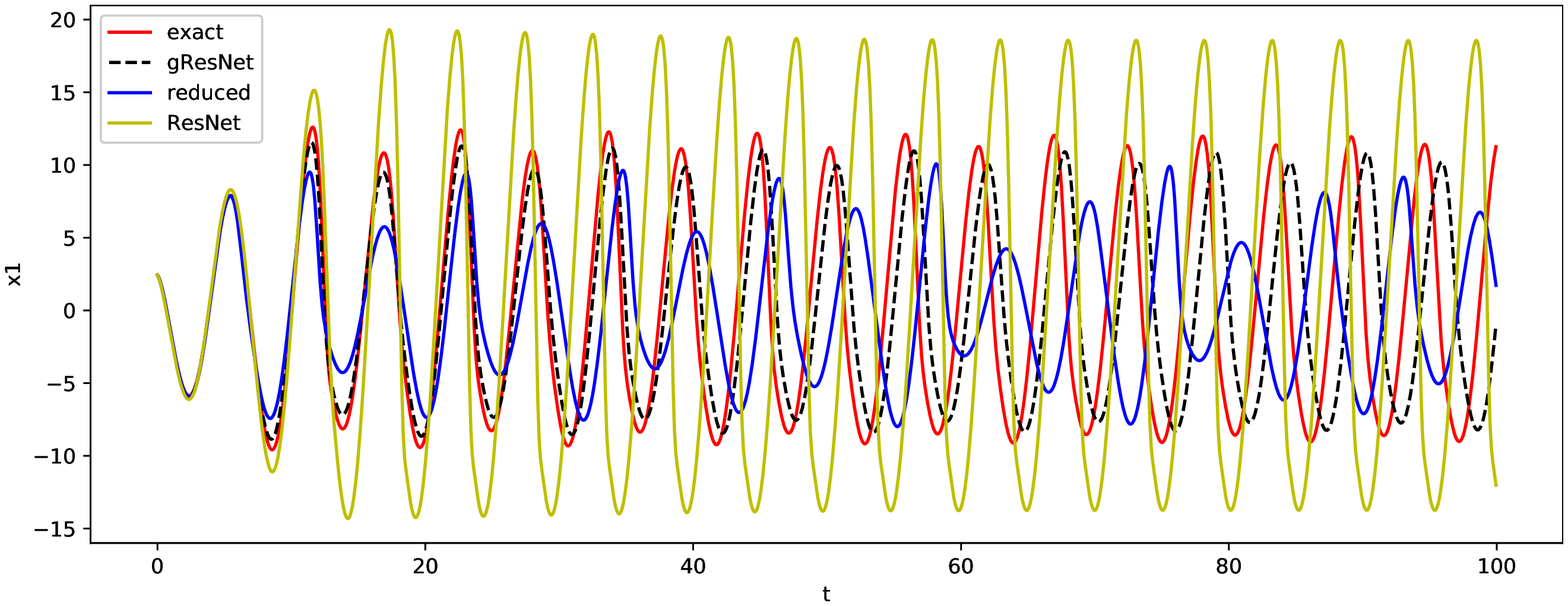}
		\includegraphics[width=12cm]{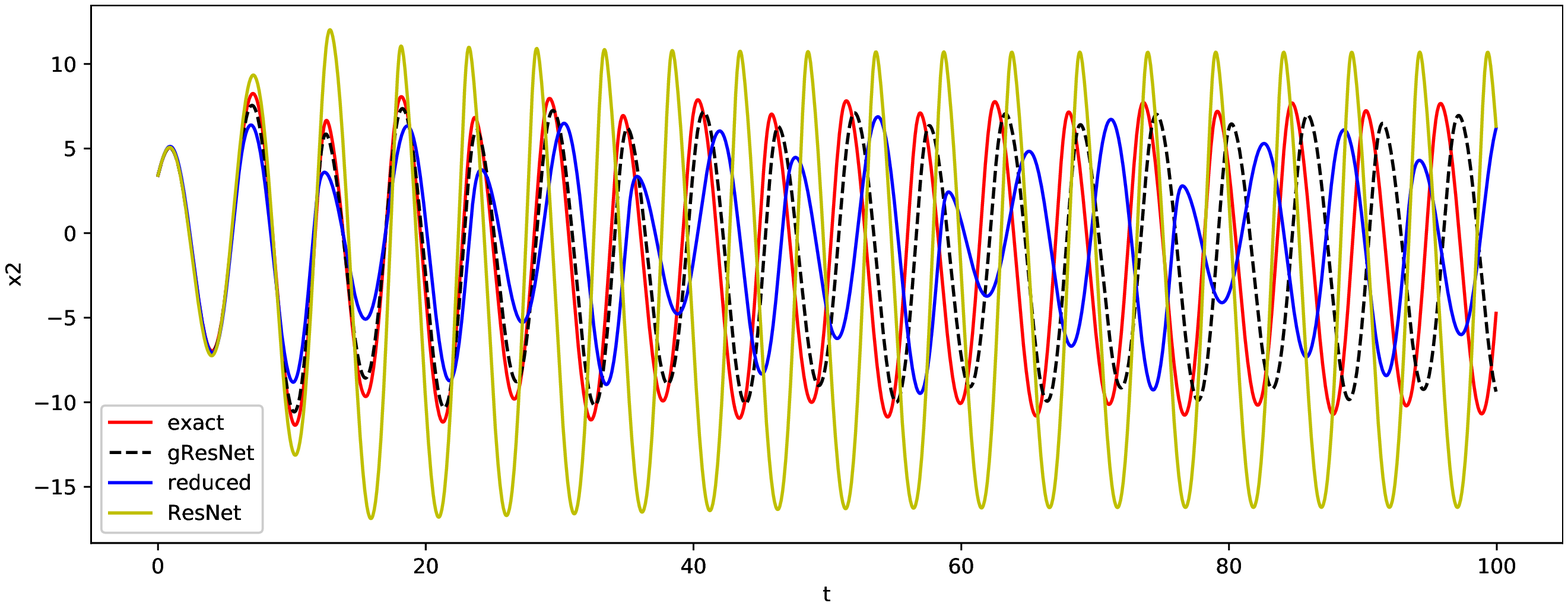}
		\includegraphics[width=12cm]{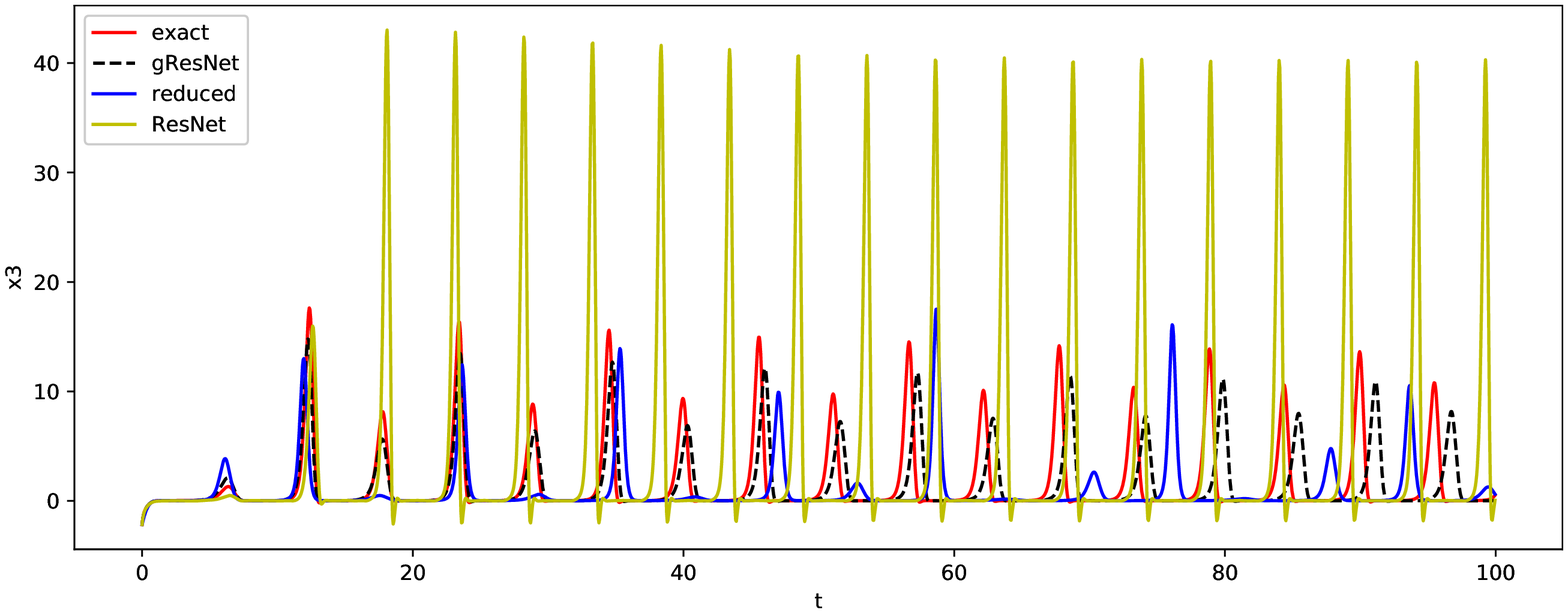}
		\includegraphics[width=12cm]{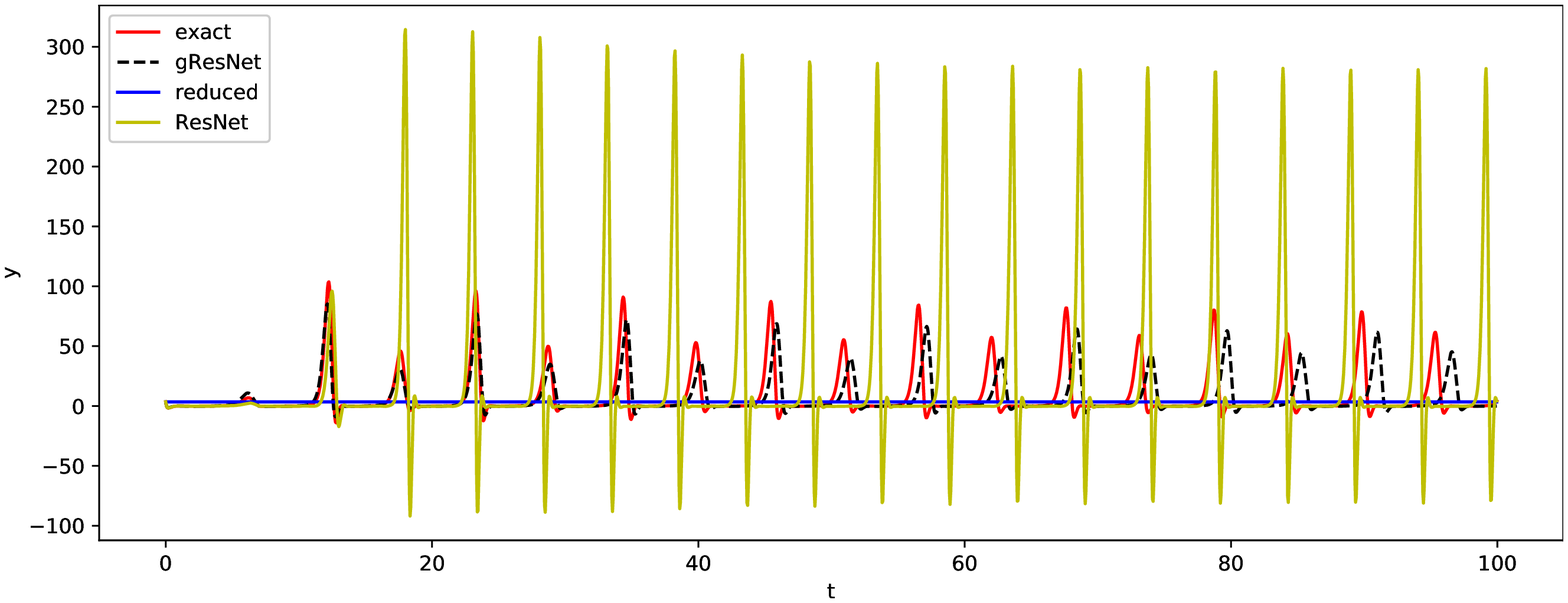}
                \caption{Example 6. Trajectory predictions by
                  different models using initial condition (2.4350451,
                  3.416925, -2.16129375, 3.4650658). Note that the
                  Reduced system does not contain variable $y$.}
		\label{fig:ex6_traj}
	\end{center}

\end{figure}

\begin{figure}[htbp]
	\begin{center}
		\includegraphics[width=12cm]{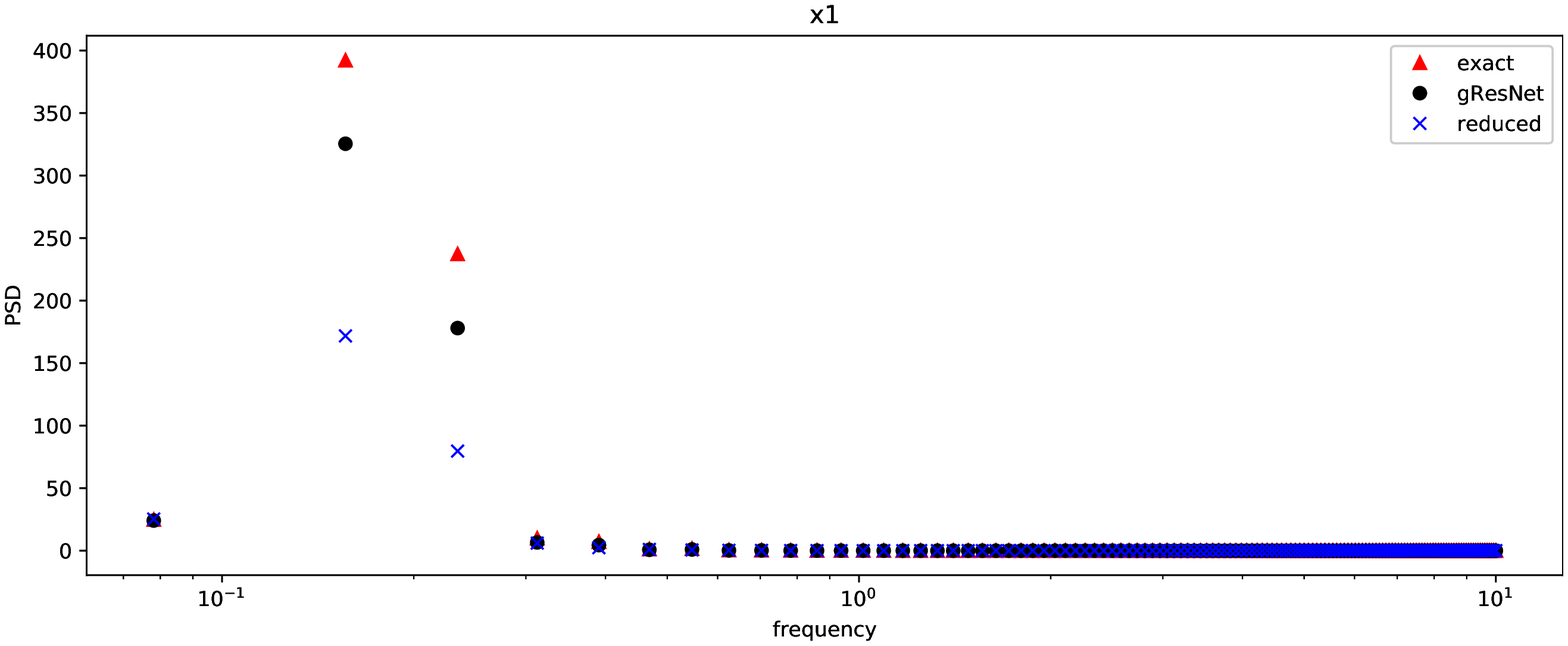}
		\includegraphics[width=12cm]{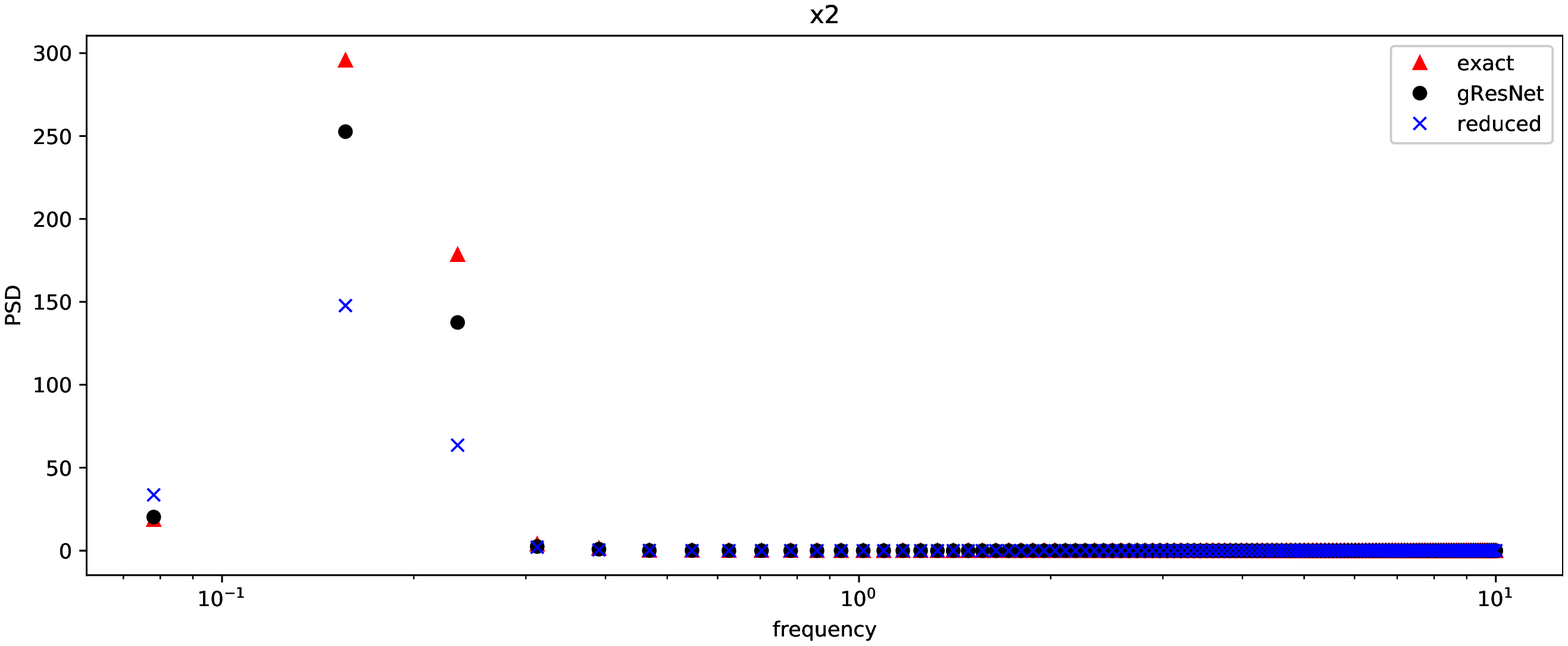}
		\includegraphics[width=12cm]{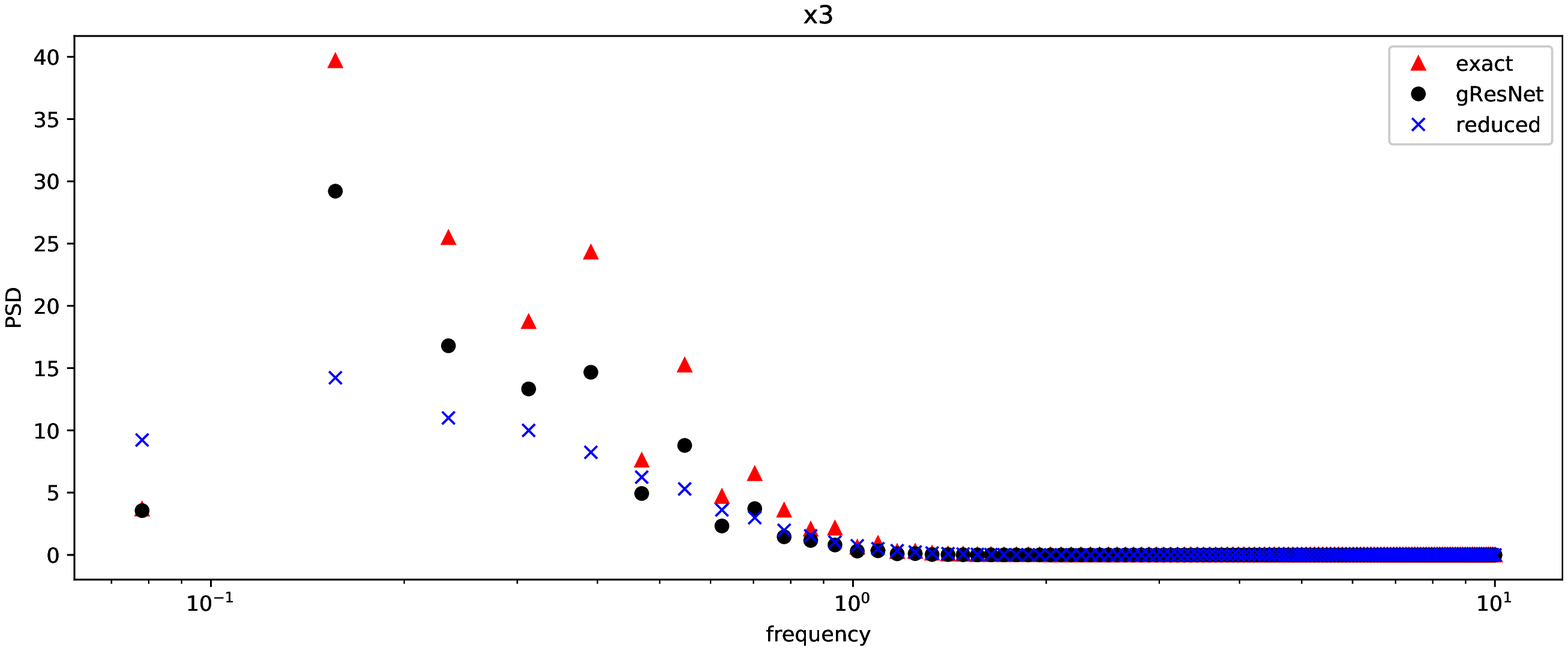}
		\includegraphics[width=12cm]{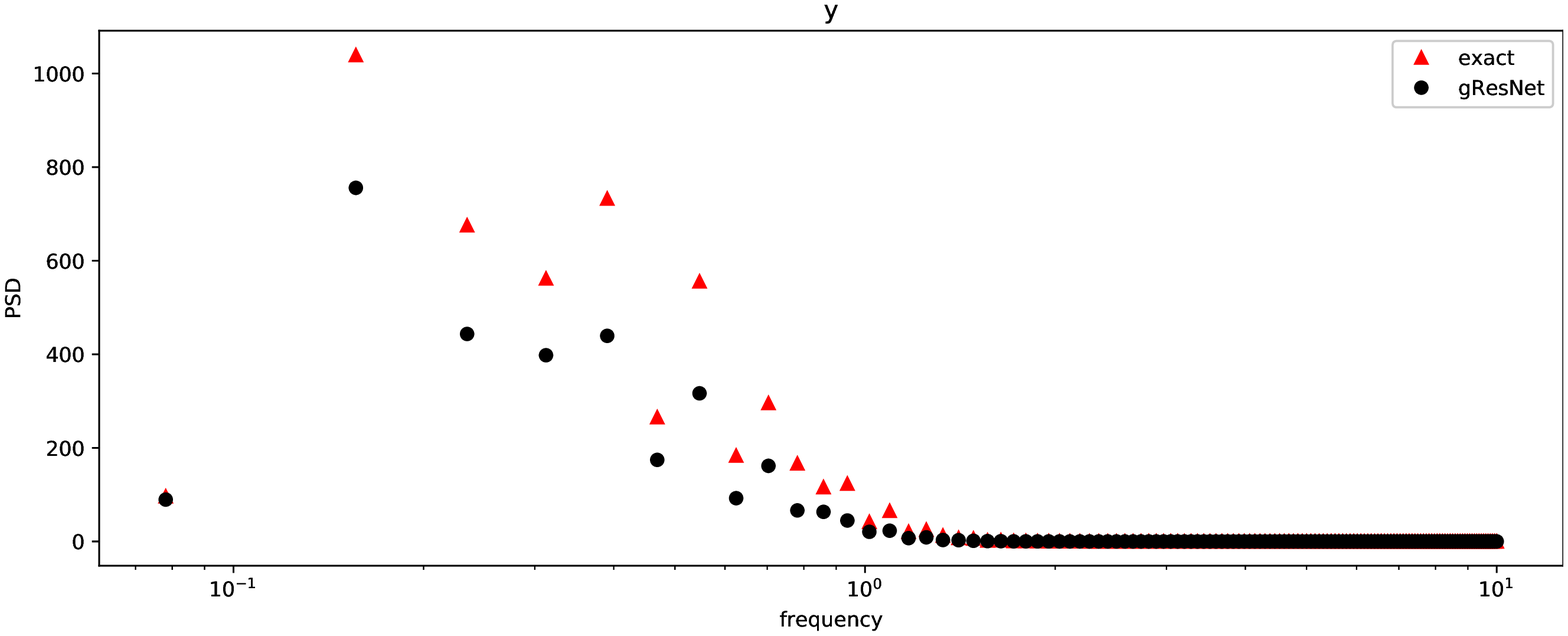}
\caption{Example 6. Power spectral density (PSD) of each trajectory
  obtained by different models.}
                \label{fig:ex6_freq}
	\end{center}
\end{figure}

The prior model \eqref{ODE3} is a good approximation of the true model
\eqref{ODE4} when the parameter $\epsilon\ll 1$. Here we set
$\epsilon = 0.1$, which is not exceedingly small. In this case, the
approximation offered by the prior model \eqref{ODE3} is relatively
coarse. We fix the computational domain as $D = [-15, 15]
\times [-15, 10] \times [-5, 25] \times [-30, 140]$ and set the time
lag as
$\Delta  = 0.05$. After satisfactory training, we utilize the trained
gResNet model to conduct long-term prediction for time up to $t=100$. The results from an arbitrarily
chosen initial condition are shown in \figref{fig:ex6_traj}. For
comparison, we also plot the prediction results obtained by the
reduced system \eqref{ODE3} (labeled as ``Reduced''), the standard
ResNet model, along with the reference exact solution via solving the
true system \eqref{ODE4} numerically.
We first observe that the gResNet method offers significantly better
results than the standard ResNet. This again confirms that it is
highly advantageous to have a good prior model. In this case, the
reduced system \eqref{ODE3} in gResNet is obviously much better than the curde model
of identity operator in the standard ResNet. More careful examination
of the results also reveals that the gResNet has better predictive
accuracy than the reduced model, especially in term of capturing the
correct phase over longer time. To visual this closely, we compute the
spectrum density of the trajectories in \figref{fig:ex6_freq}, in order to examine the dominant
frequencies in the solutions.
It can be clearly seen that the gResNet offers significant
improvement in accuracy over the reduced system.


\section{Conclusion} \label{sec:conclusions}

We presented a generalized residue network (gResNet) framework for
effective learning of unknown
governing equations from observational data. The gResNet incorporates
the standard residue network (ResNet) as a special case. In gResNet,
``residue'' is more broadly defined as the difference between the data and
the prediction of a prior model, and a deep neural network is used to
model the residue. Consequently, gResNet can be considered as a model
correction to the prior model, which is usually a reduced/coarse
model. In situations where prior models are not available, we propose
a few choices for fast construction of prior models using the same
data set and without incurring much computational cost. Various
numerical examples were presented and demonstrated that gResNet is a
viable tool for equation learning and offers better accuracy than the
standard ResNet. It is especially useful as a model correction tool,
to improve the predictive accuracy of an existing coarse model.

\bibliographystyle{siamplain}
\bibliography{LearningEqs,random}

\begin{thebibliography}{10}

\bibitem{tensorflow2015}
{\sc M.~Abadi, A.~Agarwal, P.~Barham, E.~Brevdo, Z.~Chen, C.~Citro, G.~S.
  Corrado, A.~Davis, J.~Dean, M.~Devin, S.~Ghemawat, I.~Goodfellow, A.~Harp,
  G.~Irving, M.~Isard, Y.~Jia, R.~Jozefowicz, L.~Kaiser, M.~Kudlur,
  J.~Levenberg, D.~Man\'{e}, R.~Monga, S.~Moore, D.~Murray, C.~Olah,
  M.~Schuster, J.~Shlens, B.~Steiner, I.~Sutskever, K.~Talwar, P.~Tucker,
  V.~Vanhoucke, V.~Vasudevan, F.~Vi\'{e}gas, O.~Vinyals, P.~Warden,
  M.~Wattenberg, M.~Wicke, Y.~Yu, and X.~Zheng}, {\em {TensorFlow}: Large-scale
  machine learning on heterogeneous systems}, 2015,
  \url{https://www.tensorflow.org/}.
\newblock Software available from tensorflow.org.

\bibitem{BayarriEtAl_07}
{\sc M.~Bayarri, J.~Berger, R.~Paulo, J.~Sacks, J.~Cafeo, C.~L. J.~Cavendish,
  and J.~Tu}, {\em A framework for validation of computer models},
  Technometrics, 49 (2007), pp.~138--154.

\bibitem{brunton2017chaos}
{\sc S.~L. Brunton, B.~W. Brunton, J.~L. Proctor, E.~Kaiser, and J.~N. Kutz},
  {\em Chaos as an intermittently forced linear system}, Nature Communications,
  8 (2017).

\bibitem{brunton2016discovering}
{\sc S.~L. Brunton, J.~L. Proctor, and J.~N. Kutz}, {\em Discovering governing
  equations from data by sparse identification of nonlinear dynamical systems},
  Proc. Natl. Acad. Sci. U.S.A., 113 (2016), pp.~3932--3937.

\bibitem{Chan_2018}
{\sc S.~Chan and A.~H. Elsheikh}, {\em A machine learning approach for
  efficient uncertainty quantification using multiscale methods}, Journal of
  Computational Physics, 354 (2018), p.~493–511,
  \url{https://doi.org/10.1016/j.jcp.2017.10.034},
  \url{http://dx.doi.org/10.1016/j.jcp.2017.10.034}.

\bibitem{Han8505}
{\sc J.~Han, A.~Jentzen, and W.~E}, {\em Solving high-dimensional partial
  differential equations using deep learning}, Proceedings of the National
  Academy of Sciences, 115 (2018), pp.~8505--8510,
  \url{https://doi.org/10.1073/pnas.1718942115},
  \url{https://www.pnas.org/content/115/34/8505},
  \url{https://arxiv.org/abs/https://www.pnas.org/content/115/34/8505.full.pdf}.

\bibitem{HeXiu_JCP16}
{\sc Y.~He and D.~Xiu}, {\em Numerical strategy for model correction using
  physical constraints}, J. Comput. Phys., 313 (2016), pp.~617--634.

\bibitem{HESTHAVEN201855}
{\sc J.~Hesthaven and S.~Ubbiali}, {\em Non-intrusive reduced order modeling of
  nonlinear problems using neural networks}, Journal of Computational Physics,
  363 (2018), pp.~55 -- 78,
  \url{https://doi.org/https://doi.org/10.1016/j.jcp.2018.02.037},
  \url{http://www.sciencedirect.com/science/article/pii/S0021999118301190}.

\bibitem{HigdonEtAl_04}
{\sc D.~Higdon, M.~Kennedy, J.~Cavendish, J.~Cafeo, and R.~Ryne}, {\em
  Combining field data and computer simulations for calibration and
  prediction}, SIAM J. Sci. Comput., 26 (2004), pp.~448--466.

\bibitem{JosephM_09}
{\sc V.~Joseph and S.~Melkote}, {\em Statistical adjustments to engineering
  models}, J. Quality Tech., 41 (2009), pp.~362--375.

\bibitem{kang2019ident}
{\sc S.~H. Kang, W.~Liao, and Y.~Liu}, {\em {IDENT}: Identifying differential
  equations with numerical time evolution}, arXiv preprint arXiv:1904.03538,
  (2019).

\bibitem{karumuri2019simulator}
{\sc S.~Karumuri, R.~Tripathy, I.~Bilionis, and J.~Panchal}, {\em
  Simulator-free solution of high-dimensional stochastic elliptic partial
  differential equations using deep neural networks}, arXiv preprint
  arXiv:1902.05200,  (2019).

\bibitem{KennedyOHagan01}
{\sc M.~Kennedy and A.~{O'Hagan}}, {\em Bayesian calibration of computer
  models}, J. R. Statist., 63 (2001), pp.~425--464.

\bibitem{Khoo2018}
{\sc Y.~Khoo, J.~Lu, and L.~Ying}, {\em Solving for high-dimensional committor
  functions using artificial neural networks}, Research in the Mathematical
  Sciences, 6 (2018), p.~1, \url{https://doi.org/10.1007/s40687-018-0160-2},
  \url{https://doi.org/10.1007/s40687-018-0160-2}.

\bibitem{kutz2016dynamic}
{\sc J.~N. Kutz, S.~L. Brunton, B.~W. Brunton, and J.~L. Proctor}, {\em Dynamic
  mode decomposition: data-driven modeling of complex systems}, SIAM, 2016.

\bibitem{Kutz_2014}
{\sc J.~N. Kutz, S.~L. Brunton, D.~M. Luchtenburg, C.~W. Rowley, and J.~H. Tu},
  {\em On dynamic mode decomposition: Theory and applications}, Journal of
  Computational Dynamics, 1 (2014), p.~391–421,
  \url{https://doi.org/10.3934/jcd.2014.1.391},
  \url{http://dx.doi.org/10.3934/jcd.2014.1.391}.

\bibitem{long2018pde}
{\sc Z.~Long, Y.~Lu, and B.~Dong}, {\em {PDE-Net} 2.0: Learning {PDEs} from
  data with a numeric-symbolic hybrid deep network}, arXiv preprint
  arXiv:1812.04426,  (2018).

\bibitem{long2017pde}
{\sc Z.~Long, Y.~Lu, X.~Ma, and B.~Dong}, {\em {PDE}-net: Learning {PDE}s from
  data}, in Proceedings of the 35th International Conference on Machine
  Learning, J.~Dy and A.~Krause, eds., vol.~80 of Proceedings of Machine
  Learning Research, Stockholmsm?ssan, Stockholm Sweden, 10--15 Jul 2018, PMLR,
  pp.~3208--3216.

\bibitem{Mangan20170009}
{\sc N.~M. Mangan, J.~N. Kutz, S.~L. Brunton, and J.~L. Proctor}, {\em Model
  selection for dynamical systems via sparse regression and information
  criteria}, Proceedings of the Royal Society of London A: Mathematical,
  Physical and Engineering Sciences, 473 (2017).

\bibitem{nguyen2019like}
{\sc D.~Nguyen, S.~Ouala, L.~Drumetz, and R.~Fablet}, {\em {EM}-like learning
  chaotic dynamics from noisy and partial observations}, arXiv preprint
  arXiv:1903.10335,  (2019).

\bibitem{PavliotisStuart_2008}
{\sc G.~Pavliotis and A.~Stuart}, {\em Multiscale methods: averaging and
  homogenization}, Springer, 2008.

\bibitem{Pawar_2019}
{\sc S.~Pawar, S.~M. Rahman, H.~Vaddireddy, O.~San, A.~Rasheed, and P.~Vedula},
  {\em A deep learning enabler for nonintrusive reduced order modeling of fluid
  flows}, Physics of Fluids, 31 (2019), p.~085101,
  \url{https://doi.org/10.1063/1.5113494},
  \url{http://dx.doi.org/10.1063/1.5113494}.

\bibitem{pulch2013polynomial}
{\sc R.~Pulch}, {\em Polynomial chaos for semiexplicit differential algebraic
  equations of index 1}, Int. J. Uncertain. Quantif., 3 (2013).

\bibitem{QianWu_08}
{\sc Z.~Qian and C.~Wu}, {\em Bayesian hierarchical modeling for integration
  low-accuracy and high-accuracy experiements}, Technometrics, 50 (2008),
  pp.~192--204.

\bibitem{qin2018data}
{\sc T.~Qin, K.~Wu, and D.~Xiu}, {\em Data driven governing equations
  approximation using deep neural networks}, J. Comput. Phys., 395 (2019),
  pp.~620 -- 635.

\bibitem{raissi2018deep}
{\sc M.~Raissi}, {\em Deep hidden physics models: {Deep} learning of nonlinear
  partial differential equations}, Journal of Machine Learning Research, 19
  (2018), pp.~1--24.

\bibitem{RAISSI2018125}
{\sc M.~Raissi and G.~E. Karniadakis}, {\em Hidden physics models: Machine
  learning of nonlinear partial differential equations}, Journal of
  Computational Physics, 357 (2018), pp.~125 -- 141.

\bibitem{raissi2017machine}
{\sc M.~Raissi, P.~Perdikaris, and G.~E. Karniadakis}, {\em Machine learning of
  linear differential equations using gaussian processes}, J. Comput. Phys.,
  348 (2017), pp.~683--693.

\bibitem{raissi2017physics1}
{\sc M.~Raissi, P.~Perdikaris, and G.~E. Karniadakis}, {\em Physics informed
  deep learning (part i): Data-driven solutions of nonlinear partial
  differential equations}, arXiv preprint arXiv:1711.10561,  (2017).

\bibitem{raissi2017physics2}
{\sc M.~Raissi, P.~Perdikaris, and G.~E. Karniadakis}, {\em Physics informed
  deep learning (part ii): Data-driven discovery of nonlinear partial
  differential equations}, arXiv preprint arXiv:1711.10566,  (2017).

\bibitem{raissi2018multistep}
{\sc M.~Raissi, P.~Perdikaris, and G.~E. Karniadakis}, {\em Multistep neural
  networks for data-driven discovery of nonlinear dynamical systems}, arXiv
  preprint arXiv:1801.01236,  (2018).

\bibitem{RAY2018166}
{\sc D.~Ray and J.~S. Hesthaven}, {\em An artificial neural network as a
  troubled-cell indicator}, Journal of Computational Physics, 367 (2018),
  pp.~166 -- 191,
  \url{https://doi.org/https://doi.org/10.1016/j.jcp.2018.04.029},
  \url{http://www.sciencedirect.com/science/article/pii/S0021999118302547}.

\bibitem{rudy2017data}
{\sc S.~H. Rudy, S.~L. Brunton, J.~L. Proctor, and J.~N. Kutz}, {\em
  Data-driven discovery of partial differential equations}, Science Advances, 3
  (2017), p.~e1602614.

\bibitem{rudy2018deep}
{\sc S.~H. Rudy, J.~N. Kutz, and S.~L. Brunton}, {\em Deep learning of dynamics
  and signal-noise decomposition with time-stepping constraints}, J. Comput.
  Phys., 396 (2019), pp.~483--506.

\bibitem{SargsyanNG_15}
{\sc K.~Sargsyan, H.~Najm, and R.~Ghanem}, {\em On the statistical calibration
  of physical models}, Int. J. Chem. Kinetics, DOI 10.1002/kin.20906 (2015).

\bibitem{schaeffer2017learning}
{\sc H.~Schaeffer}, {\em Learning partial differential equations via data
  discovery and sparse optimization}, Proceedings of the Royal Society of
  London A: Mathematical, Physical and Engineering Sciences, 473 (2017).

\bibitem{schaeffer2017sparse}
{\sc H.~Schaeffer and S.~G. McCalla}, {\em Sparse model selection via integral
  terms}, Phys. Rev. E, 96 (2017), p.~023302.

\bibitem{schaeffer2017extracting}
{\sc H.~Schaeffer, G.~Tran, and R.~Ward}, {\em Extracting sparse
  high-dimensional dynamics from limited data}, SIAM Journal on Applied
  Mathematics, 78 (2018), pp.~3279--3295.

\bibitem{Schmid_2010}
{\sc P.~Schmid}, {\em Dynamic mode decomposition of numerical and experimental
  data}, J. Fluid Mech., 656 (2010), pp.~5--28.

\bibitem{sun2019neupde}
{\sc Y.~Sun, L.~Zhang, and H.~Schaeffer}, {\em {NeuPDE}: Neural network based
  ordinary and partial differential equations for modeling time-dependent
  data}, arXiv preprint arXiv:1908.03190,  (2019).

\bibitem{tibshirani1996regression}
{\sc R.~Tibshirani}, {\em Regression shrinkage and selection via the lasso},
  Journal of the Royal Statistical Society. Series B (Methodological),  (1996),
  pp.~267--288.

\bibitem{tran2017exact}
{\sc G.~Tran and R.~Ward}, {\em Exact recovery of chaotic systems from highly
  corrupted data}, Multiscale Model. Simul., 15 (2017), pp.~1108--1129.

\bibitem{Tripathy_2018}
{\sc R.~K. Tripathy and I.~Bilionis}, {\em Deep uq: Learning deep neural
  network surrogate models for high dimensional uncertainty quantification},
  Journal of Computational Physics, 375 (2018), p.~565–588,
  \url{https://doi.org/10.1016/j.jcp.2018.08.036},
  \url{http://dx.doi.org/10.1016/j.jcp.2018.08.036}.

\bibitem{WANG2019289}
{\sc Q.~Wang, J.~S. Hesthaven, and D.~Ray}, {\em Non-intrusive reduced order
  modeling of unsteady flows using artificial neural networks with application
  to a combustion problem}, Journal of Computational Physics, 384 (2019),
  pp.~289 -- 307,
  \url{https://doi.org/https://doi.org/10.1016/j.jcp.2019.01.031},
  \url{http://www.sciencedirect.com/science/article/pii/S0021999119300828}.

\bibitem{WangCT_09}
{\sc S.~Wang, W.~Chen, and K.~Tsui}, {\em Bayesian validation of computer
  models}, Technometrics, 51 (2009), pp.~439--451.

\bibitem{wang2019efficient}
{\sc Y.~Wang and G.~Lin}, {\em Efficient deep learning techniques for
  multiphase flow simulation in heterogeneous porous media}, 2019,
  \url{https://arxiv.org/abs/1907.09571}.

\bibitem{WuQinXiu2019}
{\sc K.~Wu, T.~Qin, and D.~Xiu}, {\em Structure-preserving method for
  reconstructing unknown hamiltonian systems from trajectory data}, arXiv
  preprint arXiv:1905.10396,  (2019).

\bibitem{WuXiu_JCPEQ18}
{\sc K.~Wu and D.~Xiu}, {\em Numerical aspects for approximating governing
  equations using data}, J. Comput. Phys., 384 (2019), pp.~200--221.

\bibitem{Zhu_2018}
{\sc Y.~Zhu and N.~Zabaras}, {\em Bayesian deep convolutional encoder–decoder
  networks for surrogate modeling and uncertainty quantification}, Journal of
  Computational Physics, 366 (2018), p.~415–447,
  \url{https://doi.org/10.1016/j.jcp.2018.04.018},
  \url{http://dx.doi.org/10.1016/j.jcp.2018.04.018}.

\end{thebibliography}

\end{document}